\documentclass[twoside,11pt]{article}

\usepackage{jmlr2e}

\usepackage[utf8]{inputenc} %
\usepackage[T1]{fontenc}    %
\usepackage{hyperref}       %
\usepackage{url}            %
\usepackage{booktabs}       %
\usepackage{amsfonts}       %
\usepackage{nicefrac}       %
\usepackage{microtype}      %
\usepackage{xcolor}         %

\usepackage{amssymb}

\usepackage{graphicx}
\usepackage{caption}
\usepackage{subcaption}
\usepackage{amsmath}
\usepackage{enumerate}  
\usepackage{algorithm}
\usepackage{algpseudocode}

\DeclareMathOperator*{\argmax}{arg\,max}
\DeclareMathOperator*{\argmin}{arg\,min}
\newcommand{\bx}{\mathbf{x}}
\newcommand{\bv}{\mathbf{v}}

\newcommand{\by}{\mathbf{y}}
\newcommand{\bxi}{\pmb{\xi}}
\newcommand{\norm}[1]{\left\lVert#1\right\rVert}
\newcommand{\floor}[1]{\lfloor #1 \rfloor}

\newtheorem{thm}{Theorem}
\newtheorem{metathm}{Meta Theorem}
\newtheorem{cor}{Corollary}
\newtheorem{lem}{Lemma}
\newtheorem{rem}{Remark}

\ShortHeadings{Exact Dynamics of INO-PCA in High Dimensions}{Demir and Do\u{g}an}
\firstpageno{1}

\begin{document}

\title{Implicitly Normalized Online PCA: A Regularized Algorithm with Exact High-Dimensional Dynamics}

\author{\name Samet Demir\textsuperscript{1} \email sdemir20@ku.edu.tr 
\AND
       \name Zafer Do\u{g}an\textsuperscript{1,2} \email zdogan@ku.edu.tr\\
       \addr \textsuperscript{1}Machine Learning and Information Processing Group, KUIS AI Center\\ \textsuperscript{2}Department of Electrical and Electronics Engineering\\
       Ko\c{c} University\\
       İstanbul, Turkey}

\editor{}

\maketitle

\begin{abstract}%
Many online learning algorithms—including classical online PCA methods—enforce explicit normalization steps that discard the evolving norm of the parameter vector. We show that this norm can in fact encode meaningful information about the underlying statistical structure of the problem, and that exploiting this information leads to improved learning behavior. Motivated by this principle, we introduce Implicitly Normalized Online PCA (INO-PCA), an online PCA algorithm that removes the unit-norm constraint and instead allows the parameter norm to evolve dynamically through a simple regularized update. We prove that in the high-dimensional limit the joint empirical distribution of the estimate and the true component converges to a deterministic measure-valued process governed by a nonlinear PDE. This analysis reveals that the parameter norm obeys a closed-form ODE coupled with the cosine similarity, forming an internal state variable that regulates learning rate, stability, and sensitivity to signal-to-noise ratio (SNR). The resulting dynamics uncover a three-way relationship between the norm, SNR, and optimal step size, and expose a sharp phase transition in steady-state performance. Both theoretically and experimentally, we show that INO-PCA consistently outperforms Oja’s algorithm and adapts rapidly in non-stationary environments. Overall, our results demonstrate that relaxing norm constraints can be a principled and effective way to encode and exploit problem-relevant information in online learning algorithms.

\end{abstract}

\begin{keywords}
  Online learning algorithms, principal component analysis, asymptotic analysis, measure-valued process, nonlinear PDE, spiked covariance model.
\end{keywords}

\section{Introduction}

Online learning algorithms frequently impose explicit constraints on the parameters---most notably, projections onto a fixed-norm set---to ensure stability and identifiability \citep{shalev2014understanding}. Classical online PCA algorithms such as Oja’s method exemplify this design choice: after each stochastic gradient step, the iterate is rescaled to lie on the unit sphere \citep{Oja1983, kumar2024oja}. While this normalization enforces numerical stability, it also eliminates all information contained in the evolving norm of the parameter vector, i.e., the current estimate. Implicit in these methods is the assumption that the norm carries no meaningful information about the underlying statistical structure or the learning dynamics.

Recent work in modern machine learning suggests the opposite: parameter norms encode problem-dependent information that can serve as a measure of progress \citep{hu2023latent, junior2025grokking, liu2023omnigrok, nanda2023progress} or stability \citep{Li2020An, merrill2021effects}. These observations raise a natural and largely unexplored research question:

\textit{For an online learning problem that is norm-invariant, can the norm of the parameters be allowed to evolve so that it encodes problem-relevant information, and can an online learning algorithm exploit this information to achieve improved performance?}

The online PCA problem \citep{Cardot2015, greenacre2022principal, bienstock2022robust, lee2023fair, kumar2023streaming, kumar2024oja} provides an ideal setting in which to investigate this question, since it is norm-invariant and can be analyzed precisely in high-dimensional regimes \citep{wang2017scaling}. Yet despite this, the role of the parameter norm itself has remained almost entirely unexamined: the hard normalization step removes any opportunity for the algorithm to use the norm as an internal state variable reflecting information about progress, stability, or signal-to-noise ratio (SNR). Here, the SNR simply measures how strong the signal is relative to the noise in the observations, which in the spiked model corresponds to the gap between the leading eigenvalue and unity.%

Motivated by this perspective, we revisit the design of online PCA algorithms by relaxing the unit-norm constraint and allowing the iterate’s norm to evolve dynamically. This leads to a remarkably simple algorithm, which we refer to as Implicitly Normalized Online PCA (INO-PCA). Instead of performing an explicit projection step, INO-PCA arises from a regularized formulation of the PCA objective and produces the update rule
\[
\mathbf{x}_{k+1} = \mathbf{x}_k + \frac{\tau}{p}\!\left( \mathbf{y}_k \mathbf{y}_k^\top \frac{\mathbf{x}_k}{\lambda_k} - \mathbf{x}_k \right)
\qquad \textit{with} \qquad
\lambda_k = \frac{\|\mathbf{x}_k\|}{\sqrt{p}},
\]
so that the norm $\lambda_k$ evolves naturally through the data stream, where $\tau$ is the learning rate, $\mathbf{y}_k \in \mathbb{R}^p$ is the data sample at $k$-th step and $\mathbf{x}_k$ is the corresponding estimate of the leading eigenvector of the (unknown) covariance matrix—that is, the first principal component. This relaxation reveals that the iterate’s norm is not merely a nuisance variable but a meaningful quantity that can encode information about the signal, noise, and gradient dynamics.

Our analysis shows that this evolving norm regulates the effective learning rate: when the norm is large, the update step is automatically damped, and when the norm is small, it is amplified. Moreover, in the spiked covariance model, the norm converges to the leading eigenvalue, providing an internal estimate of the signal strength. These effects lead to substantially improved performance relative to classical algorithms. Empirically, INO-PCA learns as quickly as Oja’s algorithm with a large step size during the initial phase, yet achieves the stable steady-state accuracy of Oja’s method with a much smaller learning rate---a combination that no fixed-step version of Oja’s method attains.

The relaxed formulation also leads to a highly tractable analysis in high dimensions. Using tools from the mean-field theory \citep{Wang2016, wang2017scaling, Wangica2017, Wang2019, bond2024exploring}, we prove that the joint empirical distribution of the estimate and the true principal component converges to a deterministic measure-valued process governed by a nonlinear PDE. From this PDE, we derive closed-form ordinary differential equations describing the coupled evolution of the cosine similarity $Q_t$, i.e., the alignment between the current estimate at time $t$ and the true leading eigenvector, and the norm $\lambda_t$. These equations reveal a three-way interaction among the norm, the SNR parameter, and the optimal instantaneous learning rate, and they expose a sharp phase transition in steady-state recovery depending on the SNR. The resulting theory not only predicts the full learning trajectory with high accuracy but also clarifies the role of initialization, the benefits of adaptive step sizes, and the mechanism by which implicit normalization improves performance.

Beyond the PCA setting, our findings suggest that strict normalization constraints may suppress informative aspects of parameter dynamics that arise from the interaction between the data and the update rule. Allowing the norm to evolve can enrich the internal state of the algorithm in ways that improve the learning dynamics. INO-PCA illustrates this principle in its simplest form, showing that exploiting norm information can yield faster learning, more stable behavior, and better adaptation to non-stationary environments, all without increasing computational complexity.

\noindent Overall, the main contributions of this work are as follows:
\begin{itemize}
    \item We introduce \emph{Implicitly Normalized Online PCA} (INO-PCA), a simple online PCA algorithm that removes the hard unit-norm constraint and instead allows the parameter norm to evolve in a data-dependent manner.
    \item We show that this evolving norm encodes meaningful statistical information: it regulates the effective learning rate, reflects the underlying signal strength, and improves stability and convergence behavior.
    \item We provide an exact high-dimensional analysis of INO-PCA by proving that the joint empirical distribution of the estimate and true component converges to a deterministic measure-valued process governed by a nonlinear PDE.
    \item From this PDE, we derive closed-form ODEs for the cosine similarity and the evolving norm, revealing a three-way relationship among norm, SNR, and optimal learning rate, and uncovering a sharp phase transition in steady-state recovery.
    \item Empirically, we demonstrate that INO-PCA consistently outperforms classical online PCA algorithms, achieves both fast initial learning and strong steady-state accuracy, and adapts rapidly under non-stationary environments.
\end{itemize}

The rest of the paper is organized as follows: Section \ref{sec:problem-and-algorithm} describes the problem formulation and the proposed online PCA algorithm. Our asymptotical characterization of learning dynamics of the algorithm is given in Section \ref{sec:dynamics}. Experimental results (on simulation and real-world settings) are provided in \ref{sec:experimental_results}. Finally, an informal derivation of the main theoretical result is explained in Section \ref{sec:exchangebility_and_derivation} while the formal proof is detailed in the appendix.

\paragraph{Notation} 
Throughout this paper, we use lowercase non-bold letters for scalars (e.g., $\lambda,\tau$), and boldfaced lowercase letters for $p$-dimensional vectors (e.g., $\bx$). The $i$-th element of a vector is shown by superscript $i$ (e.g., $x^i$). The Euclidean-norm of a vector is denoted by $\norm{.}$ (e.g., $\norm{\bx}$). The subscripts $k$ and $t$ of quantities denote the discrete-time iteration step (e.g., $\bx_k$) and the continuous-time step (e.g., $Q_t$), respectively. The subscript $s$ is used to indicate the steady-state (e.g., $Q_{s} = \lim_{t \to \infty} Q_t$). The big-O notation, denoted by $\mathcal{O}(\cdot)$, is employed to provide an upper bound on the growth rate of a function.

\section{Setting and the proposed algorithm}
\label{sec:problem-and-algorithm}

In this section, we start with a description of the problem formulation for our theoretical setting, then explain the proposed algorithm that utilizes the norm to achieve improved learning dynamics, and finally, discuss how the proposed algorithm can be analyzed in high-dimensions while transitioning to the next section, which provides our theoretical characterization.

\subsection{Problem formulation}
\label{sec:problem-formulation}

A fundamental theoretical model for studying principal component estimation in high dimensions is the \emph{spiked covariance model} \citep{Johnstone2001, mergny2024spectral}, which offers a clean and analytically tractable framework for understanding the statistical and dynamical behavior of PCA algorithms. In this model, each observation is generated as    
\begin{equation}
    \mathbf{y}_k = \sqrt{\frac{\omega}{p}}\, c_k\, \pmb{\xi} + \mathbf{a}_k,
    \label{eq:data_model}
\end{equation}
where \( \pmb{\xi} \in \mathbb{R}^p \) is the true leading eigenvector, \( c_k \sim \mathcal{N}(0,1) \) is a one-dimensional latent signal component, and \( \mathbf{a}_k \sim \mathcal{N}(0, \mathbf{I}) \) represents isotropic noise. The parameter \( \omega > 0 \) controls the relative magnitude of the signal and the ambient noise and therefore serves as the signal-to-noise ratio (SNR), determining the eigengap between the leading eigenvalue \( 1+\omega \) and the bulk eigenvalue \( 1 \). We adopt the normalization \( \|\pmb{\xi}\| = \sqrt{p} \), ensuring that the entries of \( \pmb{\xi} \) remain \( \mathcal{O}(1) \) as \( p \to \infty \). Within this setting, our goal is to estimate the leading principal component \( \pmb{\xi} \) in an \emph{online} fashion, processing each sample \( \mathbf{y}_k \) exactly once.

A classical approach to this problem is \emph{Oja’s algorithm} \citep{Oja1983}, which performs a stochastic gradient descent (SGD) step followed by explicit normalization:
\begin{align}
    \hat{\mathbf{x}}_{k+1} &= \mathbf{x}_k + \frac{\tau}{p}\, \mathbf{y}_k \mathbf{y}_k^\top \mathbf{x}_k, \label{eq:oja_gradient_step}\\
    \mathbf{x}_{k+1} &= \frac{\sqrt{p}\, \hat{\mathbf{x}}_{k+1}}{\|\hat{\mathbf{x}}_{k+1}\|}. \label{eq:oja_projection_step}
\end{align}
The normalization step in \eqref{eq:oja_projection_step} is essential for stability but obscures the potential role of norm in the update dynamics. However, the norm can encode problem-dependent information (such as signal strength) that can be utilized to improve learning dynamics. This motivates alternative formulations that achieve \emph{implicit normalization} through regularization rather than projection.

\subsection{Proposed algorithm}
We introduce \emph{Implicitly Normalized Online PCA (INO-PCA)}, a regularized online algorithm derived from the optimization problem
\begin{equation}
    \hat{\mathbf{x}} = \argmin_{\mathbf{x}} \left( -\tfrac{1}{2}\mathbf{x}^\top \mathbf{\Sigma}\mathbf{x} + \tfrac{\eta}{l}\|\mathbf{x}\|^l \right),
    \label{eq:formulation}
\end{equation}
where \( \mathbf{\Sigma} \) is the population covariance, \( \eta > 0 \) controls the strength of regularization, and \( l > 2 \) determines the degree of the penalty. The higher-order norm penalty acts as a \emph{soft constraint} on the norm of \(\mathbf{x}\), replacing the hard normalization in Oja’s rule and giving rise to the implicit normalization characteristic of INO-PCA. Choosing \(\eta = \Theta(p^{-(l-2)/2})\) ensures the penalty term remains balanced in the high-dimensional limit.

Focusing on the cubic regularization case \( l = 3 \) with \( \eta = 1/\sqrt{p} \), we first obtain the following online update rule:
\begin{equation}
    \mathbf{x}_{k+1} = \mathbf{x}_{k} + \frac{\tau}{p}\!\left( \mathbf{y}_k \mathbf{y}_k^\top \mathbf{x}_{k} - \lambda_k \mathbf{x}_{k} \right),
    \label{eq:regularized_update}
\end{equation}
where \( \lambda_k = \|\mathbf{x}_k\|/\sqrt{p} \) serves as a \emph{self-normalizing scale factor}. By relaxing the constraint and allowing the iterate’s norm to evolve dynamically, this update rule preserves the essential directional learning while avoiding the abrupt rescaling inherent to projection-based methods, such as Oja's algorithm. The evolution of the norm $\lambda_k$ is regulated by the higher-order penalty in \eqref{eq:formulation}, which induces a shrinkage term proportional to \( \lambda_k \mathbf{x}_k \). In effect, the update maintains a balance between the signal-amplifying term \( \mathbf{y}_k \mathbf{y}_k^\top \mathbf{x}_k \) and the regularizing shrinkage, ensuring that the norm remains stable without explicit normalization. As shown in Appendix~\ref{appendix:equivalence_to_oja}, the learning dynamics induced by this update coincide with those of Oja’s algorithm at the level of the limiting ODE for the cosine similarity $Q_t$, despite the absence of explicit normalization. Importantly, although \eqref{eq:regularized_update} resembles Oja’s algorithm in performance, it differs in a key structural aspect: the norm is allowed to evolve and directly participates in the update dynamics. This evolution reveals information about the underlying signal strength, which is suppressed in classical normalization-based methods.

Having obtained an update rule with a dynamically evolving norm that nonetheless matches the learning dynamics of classical online PCA (specifically Oja’s method), we can leverage the problem-dependent information encoded in the norm to further stabilize learning and accelerate convergence. Our proposed algorithm, INO-PCA, incorporates this idea by scaling the gradient direction by \( 1/\lambda_k \), yielding the following update:
\begin{equation}
    \mathbf{x}_{k+1} = \mathbf{x}_k + \frac{\tau}{p}\!\left( \mathbf{y}_k \mathbf{y}_k^\top \frac{\mathbf{x}_k}{\lambda_k} - \mathbf{x}_k \right).
    \label{eq:dynamics_update}
\end{equation}
The intuition, verified explicitly in our setting, is that if the initial norm $\lambda_0$ is less than the leading eigenvalue, then $\lambda_k$ monotonically increases and converges to the leading eigenvalue in the steady state. Thus, it can serve as an internal estimator of signal strength and a measure of progress. We further show that \( \lambda_k \) remains bounded throughout the dynamics, regardless of whether its initialization is above or below the steady-state value (see Appendix~\ref{appendix:bounded_lambda_k} for the proof and Figure \ref{fig:lambda0} for an illustration), ensuring that the update rule remains stable. Since \( \lambda_k \) provides a reliable measure of both progress and signal-to-noise conditions, scaling the gradient inversely by \( \lambda_k \) stabilizes the dynamics as the learning proceeds while allowing a high effective learning rate initially, thereby accelerating the learning. This mechanism induces an intrinsic coupling between the update direction and the evolving norm, allowing INO-PCA to automatically regulate its effective step size in response to the data stream.

Overall, the INO-PCA update \eqref{eq:dynamics_update} can be viewed as a \emph{regularized stochastic gradient method} that preserves the essential structure of Oja’s rule while leveraging a dynamically evolving norm to encode problem-specific information. Before proceeding to the theoretical analysis, we conclude this section with two remarks: one clarifying the structural distinction between INO-PCA and other online PCA algorithms, and another describing its natural extension to the multi-component setting.

\begin{rem}[Distinction from other algorithms without explicit normalization]$ $\\
    There exist other online PCA algorithms, such as  Krasulina's method \citep{Krasulina1969, Balsubramani2013}, that do not involve explicit normalization of the estimates. Yet, the distinct advantage of our technique is that it explicitly utilizes the norm to achieve improved learning dynamics by design. For example, Krasulina's method allows norm drift, but in Krasulina’s update, the gradient term is orthogonal to the iterate by construction; consequently, the norm drift is incidental and carries no statistical information. In contrast, INO-PCA intentionally couples the update direction with the current norm, making $\lambda_k$ an informative and dynamically meaningful scalar state. Note that since Oja's algorithm and Krasulina's method are shown to be identical to within second-order terms, we do not explicitly compare against Krasulina's method, but our comparison with Oja's algorithm is also applicable (in most cases) for a comparison with Krasulina's method as well.
\end{rem}

\begin{rem}[Extension to multiple principal components] $ $\\
    Although our theoretical analysis focuses on recovering the leading principal component, INO-PCA naturally generalizes to multiple components via orthogonalization (see Appendix~\ref{appendix:extention_to_multiplePC}).
\end{rem}

\subsection{Transition to theoretical analysis}

A key advantage of the INO-PCA update rule in \eqref{eq:dynamics_update} is that its implicit normalization and regularized gradient structure lead to a remarkably tractable description of its stochastic dynamics in high dimensions. In particular, as the ambient dimension \(p\) tends to infinity, the joint empirical distribution of the current estimate and the true eigenvector exhibits a law-of-large-numbers effect: it converges weakly to a deterministic measure-valued process. This limiting process satisfies a nonlinear partial differential equation (PDE) that exactly captures the macroscopic evolution of INO-PCA. Crucially, the resulting limiting PDE explicitly tracks the evolution of the norm $\lambda_k$, which becomes an informative macroscopic state variable. This contrasts with normalization-based updates, whose limiting dynamics collapse onto the surface of a sphere and omit norm information entirely.

From this PDE, we derive closed-form evolution equations for key performance quantities, including a scalar ordinary differential equation (ODE) governing the cosine similarity between the estimate and the true component. The resulting dynamics reveal a nontrivial coupling between the learning rate, the evolving norm, and the rate of alignment with the signal direction. This characterization allows us to identify optimal step sizes, understand how regularization affects long-term behavior, and uncover a sharp phase transition in steady-state performance as a function of the signal-to-noise ratio. In the following section, we formalize this high-dimensional limit and develop the resulting theory.

\section{Main theoretical results: learning dynamics in high dimensions} %
\label{sec:dynamics}

We analyze the dynamics of the update rule \eqref{eq:dynamics_update} in the high-dimensional scaling regime as \( p \to \infty \). Our goal is to characterize the evolution of the algorithm through a suitable representation. To this end, we define the joint empirical measure of the iterate and the true eigenvector at iteration \( k \) as the central object of our analysis:
\begin{equation}
    \mu_{k}^p(x, \xi)  \stackrel{\text { def }}{=} \frac{1}{p} \sum_{i=1}^{p} \delta\left(x-x_{k}^{i}, \xi - \xi^i \right),
    \label{eq:measure}
\end{equation}
where \( x_{k}^{i} \) and \( \xi^{i} \) denote the \( i \)-th components of the corresponding vectors. The object \( \mu_{k}^p \) is a random element of \( \mathcal{M}(\mathbb{R}^2) \), the space of probability measures on \( \mathbb{R}^2 \). Consequently, the sequence \( \{\mu_{k}^p\}_{k \ge 0} \) forms a measure-valued stochastic process.

The empirical measure provides a convenient representation for evaluating performance metrics, many of which can be expressed as functionals of \( \mu_{k}^p \). For any test function \( f : \mathbb{R}^2 \to \mathbb{R} \), we denote the integration of \( f \) against a measure \( \mu \) by
\begin{equation}
    \left\langle f, \mu \right\rangle
    \stackrel{\text{def}}{=}
    \iint_{\mathbb{R}^2} f(x, \xi) \, \mu(x, \xi)\, dx\, d\xi,
\end{equation}
which will be used extensively to express quantities such as the iterate norm, cosine similarity, and other observables derived from the joint distribution of \( (x_k^i, \xi^i) \).

To analyze the scaling limit of \( \mu_k^p \), we embed the discrete-time sequence into continuous time via the rescaling
\begin{equation}
    \mu_t(x,\xi)
    \stackrel{\text{def}}{=}
    \mu_{\lfloor pt \rfloor}^p(x,\xi),
    \label{eq:continous_embedding}
\end{equation}
where \( \lfloor \cdot \rfloor \) denotes the floor function. This choice of time rescaling is natural: each update incorporates a single data sample, so \( \Theta(p) \) iterations correspond to one effective unit of macroscopic time. By construction, \( \mu_t(x,\xi) \) is a piecewise-constant c\`adl\`ag process taking values in \( \mathcal{M}(\mathbb{R}^2) \). Since the empirical measures are random, the trajectory \( t \mapsto \mu_t \) is a random element of the Skorokhod space \( \mathcal{D}(\mathbb{R}^+, \mathcal{M}(\mathbb{R}^2)) \), in which the notion of weak convergence is well defined \citep{Kallenberg2002}.

Our main result establishes that, as \( p \to \infty \) under this time rescaling, the sequence of joint empirical measures \( \{\mu_{k}^p(x, \xi)\}_{k \ge 0} \) converges weakly to a deterministic measure-valued process \( \mu_t(x,\xi) \). Furthermore, this limit is characterized as the unique solution to a nonlinear partial differential equation (PDE) describing the evolution of the joint density of \( (x_t, \xi) \). When the PDE admits a density-valued solution, it can be solved numerically to track the evolution of the distribution over time, yielding precise predictions for the macroscopic behavior of the algorithm.

\begin{thm}\label{theo:main_conv}
Suppose the initial empirical measure \( \mu_0^p(x,\xi) \) converges weakly to a deterministic measure \( \mu_0 \in \mathcal{M}(\mathbb{R}^2) \) as \( p \to \infty \). Assume that the initial norm parameter satisfies \( \lambda_0 = \Theta(1) \), and that the initial cosine similarity between the estimate and the true leading eigenvector is nonzero, i.e., \( Q_0 \neq 0 \). Then, as \( p \to \infty \), the measure-valued stochastic process \( \{\mu_k^p\}_{k \ge 0} \) associated with the update rule \eqref{eq:dynamics_update} converges weakly to a deterministic measure-valued process \( \mu_t \).

Moreover, the limiting process \( \mu_t(x,\xi) \) is the unique solution to the following nonlinear PDE in weak form: for every positive, bounded, and \( C^3 \) test function \( f : \mathbb{R}^2 \to \mathbb{R} \),
\begin{align}\label{eq:weak_pde}
    \left<f,\mu_{t}\right>-\left<f,\mu_{0}\right> &= \int_0^t\left<G({x}, \lambda, {\xi}, Q) \frac{\partial}{\partial x}f, \mu_{\hat{t}} \right>d\hat{t} +\frac{1}{2}\int_0^t\left<J(Q) \frac{\partial^2}{\partial x^2} f ,\mu_{\hat{t}} \right>d\hat{t},\\\nonumber
\end{align}
where the drift and diffusion coefficients are given by
\begin{equation}
    G(x, \lambda, {\xi}, Q) = \tau(\omega Q {\xi} +  \frac{{x}}{\lambda} - {x}), \quad J(Q) = \tau^2(\omega Q^2 + 1),
\end{equation}
and where the macroscopic order parameters are
\begin{align}\label{eq:tracked_par}
    Q_{t}=\iint_{\mathbb{R}^2} \frac{x \xi}{\lambda_t} \mu_{t}(x, \xi) dx d\xi, \quad \textit{and} \quad
    \lambda_{t}= \sqrt{\iint_{\mathbb{R}^2} x^2 \mu_{t}(x, \xi) dx d\xi}.
\end{align}

\begin{proof}
Our analysis relies on an exchangeability assumption, which we verify in our setting. We then derive the weak-form PDE \eqref{eq:weak_pde}. These steps are described in detail in Section~\ref{sec:exchangebility_and_derivation}. For a formal proof,  we refer to Appendix~\ref{appendix:formal_proof}.
\end{proof}
\end{thm}
\vspace{-1em}
Below, we first provide two remarks discussing the nature and implications of the given Theorem, and then we provide two corollaries characterizing the time-evolution (dynamics) of the cosine-similarity $Q_t$ and the norm $\lambda_t$.  
\begin{rem}
    Online SGD algorithms for non-convex optimization problems (e.g., online PCA) are known to experience learning dynamics with two phases \citep{benarous2021searchphase}: 1) a "search" phase where the algorithm is considered to be wandering in a non-convex landscape and 2) a learning phase where the performance quickly approaches a local optimum. Here, our Theorem \ref{theo:main_conv} fully captures the dynamics in the second phase (learning) where $Q_0 \neq 0$, whereas the "search" phase behavior of the algorithm is expected to be the same as that of Oja's algorithm (with a proper rescaling of the learning rate), characterized by \citep{benarous2021searchphase}.
\end{rem}

\begin{rem}
If a density-valued solution exists, the PDE admits the following strong form:
\begin{equation}
    \frac{d}{d t} P_t(x \mid \xi) = -\frac{\partial}{\partial {x}} [G({x}, \lambda_t, {\xi}, Q_t) P_t(x \mid \xi)] + \frac{1}{2} J(Q_t) \frac{\partial^2}{\partial x^2} P_t(x \mid \xi)
    \label{eq:strong_pde}
\end{equation}
where $P_t(x \mid \xi)$ is the conditional probability density of $x$ given $\xi$ at time t.
\end{rem}

\begin{cor}
\label{cor:Qt}
Based on the weak-form PDE \eqref{eq:weak_pde}, the time evolution of the cosine similarity \( Q_t \) satisfies the following ODE:
\begin{equation}
    \frac{d}{d t} Q_{t} = \frac{\tau Q_t}{\lambda_t} (\omega - \omega Q_t^2  - \frac{\tau(\omega Q_t^2 + 1)}{2\lambda_t}).
    \label{eq:Q_t}
\end{equation}
\end{cor}

\begin{cor}
\label{cor:lambdat}
Similarly, the evolution of the norm parameter \( \lambda_t \), which controls the scale of the estimate, is governed by the ODE
\begin{equation}
    \frac{d}{d t} \lambda_t = \tau(\omega Q_t^2 + 1 - \lambda_t + \frac{\tau(\omega Q_t^2 + 1)}{2\lambda_t}).
    \label{eq:lambda_t}
\end{equation}
Proofs for Corollaries \ref{cor:Qt} and \ref{cor:lambdat} are provided in Appendix \ref{appendix:proof_of_corollaries}.
\end{cor}
Note that the ODEs in \eqref{eq:Q_t} and \eqref{eq:lambda_t} are coupled and must be solved jointly. Here, we would like to highlight that the coupled nature of the cosine similarity $Q_{t}$ and the norm $\lambda_t$ indicates that the norm $\lambda_t$ is an important state variable for our algorithm \eqref{eq:dynamics_update} in comparison to other algorithms like Oja's algorithm. In the next section, we show that the solutions of these ODEs and the PDE \eqref{eq:strong_pde} accurately predict the empirical behavior of the algorithm while comparing our algorithm (INO-PCA) with other relevant algorithms (e.g., Oja's method) in various scenarios.

\section{Experimental results} \label{sec:experimental_results}
In this section, we present our numerical results alongside additional theoretical insights. We begin by describing the experimental setup used throughout our simulations. We then demonstrate that the trajectories predicted by our high-dimensional theory closely match the empirical behavior of the algorithm. Next, we analyze the steady-state properties of INO-PCA and show that the cosine similarity exhibits a phase transition as a function of the signal-to-noise ratio \( \omega \). We also introduce an adaptive variant of the algorithm and examine its performance. In addition, we investigate the role of the initialization by varying the initial norm parameter \( \lambda_0 \) and studying its impact on the evolution of the cosine similarity. Finally, we compare INO-PCA with several related online PCA algorithms on various scenarios, including a real-world subspace learning problem on the Olivetti Faces dataset \citep{ATT_FaceDatabase}.

\subsection{Setting}
\paragraph{Initialization}
In our numerical experiments, we consider two distinct initialization schemes for the initial estimate \( \mathbf{x}_0 \). The first scheme, referred to as the \emph{cold start}, draws each coordinate independently from a standard normal distribution, i.e., \( x_0^i \sim \mathcal{N}(0,1) \). This initialization produces an estimate that is essentially uninformative about the true principal direction, leading to a cosine similarity near zero at \( t = 0 \).

The second scheme, referred to as the \emph{warm start}, initializes \( \mathbf{x}_0 \) so that the expected initial cosine similarity satisfies \( \mathbb{E}[Q_0] = c \) for some constant \( c > 0 \). The specific value of \( c \) is not essential; any modest positive alignment suffices to break the sign symmetry of the problem and avoid the unstable regime near \( Q_0 = 0 \). In our numerical experiments we select \( c = 0.1 \), following common practice in the streaming PCA literature \citep{Wangica2017, Wang2016, wang2017scaling}. Warm starts are especially useful when theoretical guarantees require nonnegative initial alignment or when one aims to reduce early-stage variance in empirical evaluations. In all figures, the initialization type can be inferred from the cosine similarity at time \( t = 0 \): values near zero indicate cold starts\footnote{Note that cold starts exhibit higher variance because the estimate begins with nearly zero alignment, placing it in the search phase as described in Remark 3. To maintain visual clarity, we therefore display error bars corresponding to one-third standard deviation.}, whereas positive values indicate warm starts.

For both initialization schemes, we scale the initial vector so that \( \| \mathbf{x}_0 \| = \sqrt{p}\, \lambda_0 \). Unless stated otherwise, we set \( \lambda_0 = 1 \), which ensures that the estimate begins with a normalized initialization.

\paragraph{Default parameters}
The following parameter values are used in all numerical results unless otherwise specified. The ambient dimension is fixed at \( p = 10{,}000 \), providing a regime where high-dimensional asymptotics offer accurate predictions. The step size is set to \( \tau = 0.5 \), balancing stability and convergence speed, and the signal-to-noise ratio parameter is chosen as \( \omega = 1 \). To obtain reliable estimates and smooth empirical curves, each figure is generated using \( 20 \) independent Monte Carlo trials.

\subsection{Theory vs. simulations}
In this subsection, we compare our theoretical predictions with numerical simulations for two illustrative examples.
\begin{figure}[htb]
    \centering
    \includegraphics[width=0.9\textwidth]{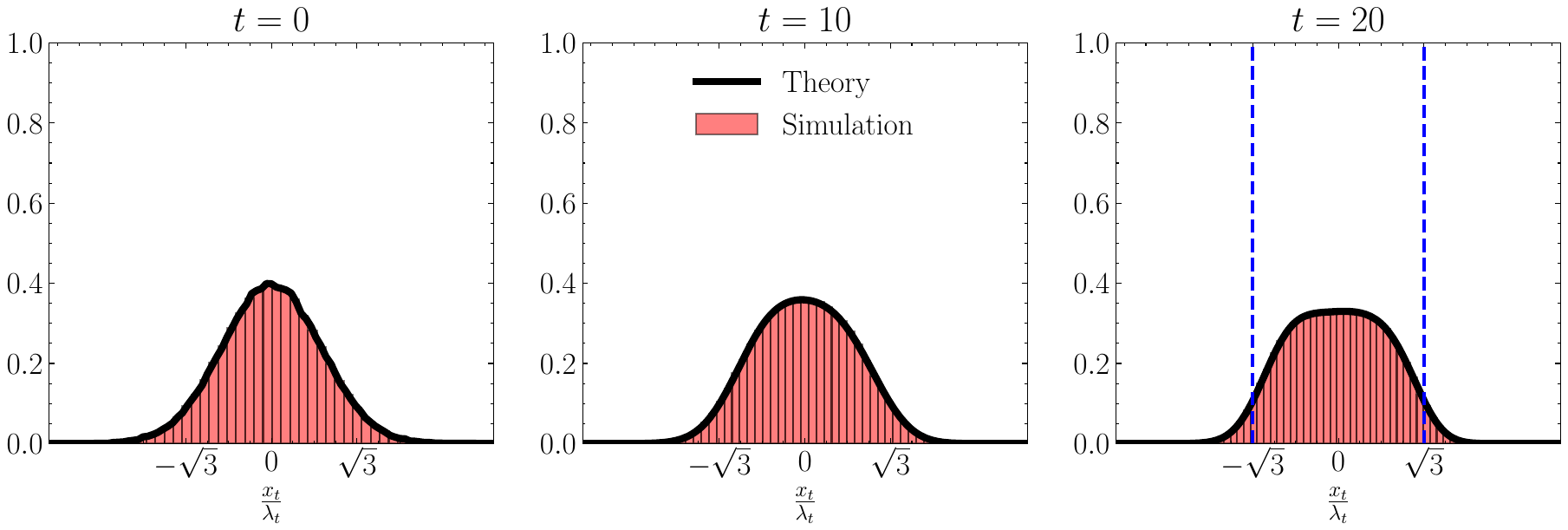}
    \caption{Theory vs.\ simulations: Comparison between the limiting asymptotic densities and the empirical densities (of $x_t/\lambda_t$) obtained from Monte Carlo simulations at different times \( t \), indicated above each panel. The vector \( \xi \) is drawn from a uniform distribution. See Example~1 for details.}
    \label{fig:density_evolution}
\end{figure}

\begin{figure}[htb]
    \centering
    \includegraphics[width=0.9\textwidth]{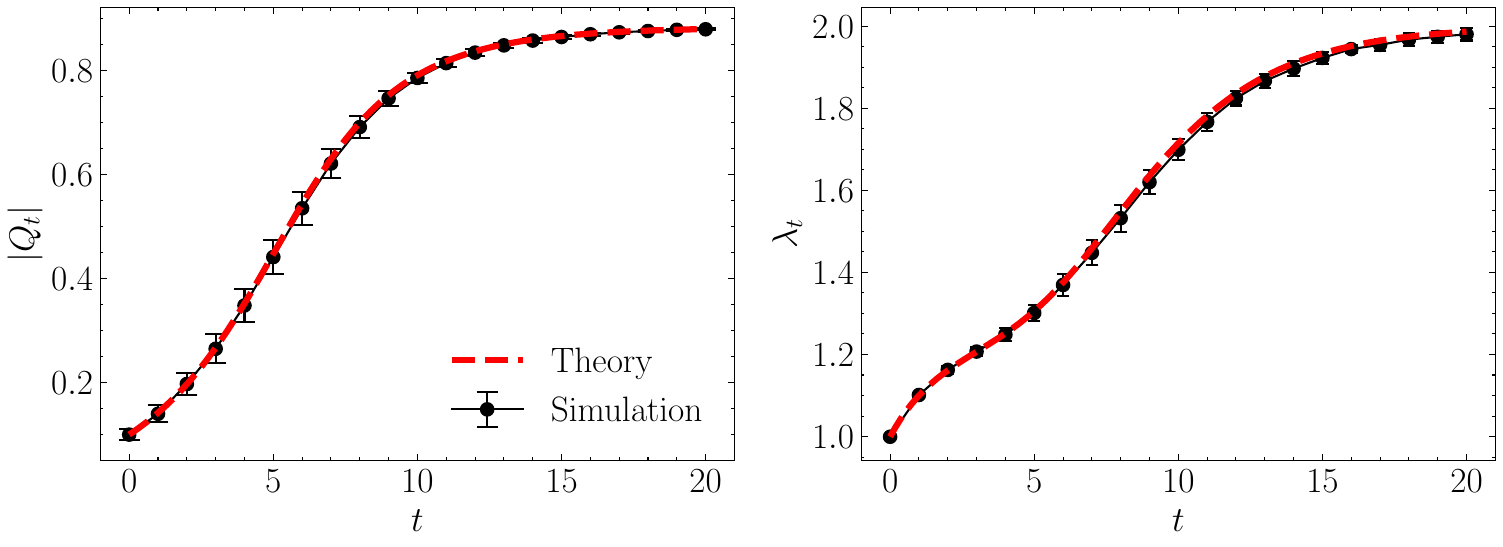}
    \caption{Theory vs.\ simulations: Evolution of \( Q_t \) (left) and \( \lambda_t \) (right) for Example~1. Solid lines correspond to the theoretical ODE predictions \eqref{eq:Q_t} and \eqref{eq:lambda_t}, respectively. The Monte Carlo estimates show the empirical mean, with bars indicating one standard deviation.}
    \label{fig:cos_lambda}
\end{figure}

\paragraph{Example 1} %
We generate the vector of interest \( \boldsymbol{\xi} \) such that its entries are sampled independently from a uniform distribution over \( [-\sqrt{3}, \sqrt{3}] \).

In Figure~\ref{fig:density_evolution}, we compare the asymptotic density $P_t(x) = \int_{\mathbb{R}} P_t(x \mid \xi)\, P(\xi)\, d\xi$ with the empirical densities obtained from simulations at three different times. The PDE \eqref{eq:strong_pde} is solved numerically to obtain the limiting conditional densities \( P_t(x \mid \xi) \). As shown in the figure, the theoretical densities closely match the empirical distributions, indicating that the PDE precisely characterizes the evolution of the distribution of $x_t$ (i.e., the distribution of the elements of the estimate at time $t$). 

In Figure~\ref{fig:cos_lambda}, we evaluate the accuracy of the ODE predictions \eqref{eq:Q_t} for \( Q_t \) and \eqref{eq:lambda_t} for \( \lambda_t \) in the setting of Example~1. The results demonstrate that the theoretical dynamics provide accurate predictions for both quantities. In particular, observe the evolution of the norm \( \lambda_t \), which increases as the learning proceeds and converges to the SNR value (the leading eigenvalue) \( \omega + 1 \), as expected and mentioned when introducing our algorithm. This also confirms our claim that the norm can encode useful information regarding the learning progress and the SNR value. Below, we demonstrate how our algorithm (INO-PCA) utilizes the information in the norm to stabilize and accelerate learning when discussing the steady-state analysis, phase transition, and comparison with algorithms.

Before switching to steady-state analysis and comparison, we would like to illustrate another example of how the found PDE asymptotically captures the evolution of the empirical densities in a case where elements of $\boldsymbol{\xi}$ are sampled from a distribution with a non-zero mean.

\begin{figure}[htb]
    \centering
    \includegraphics[width=0.9\textwidth]{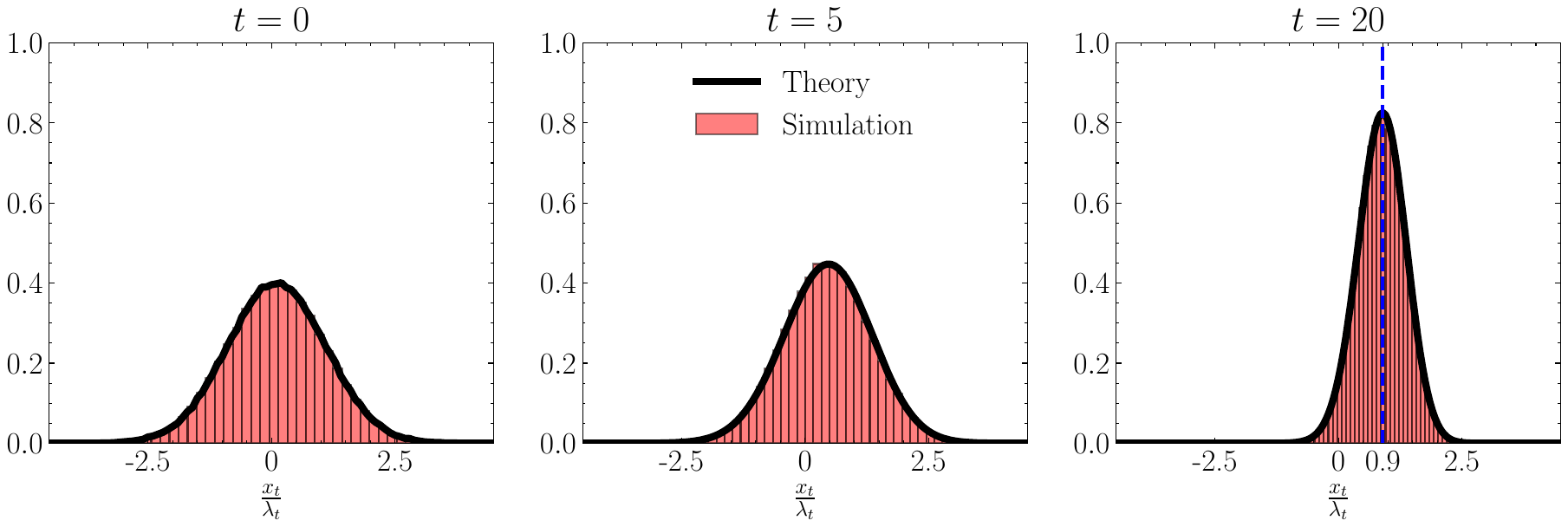}
    \caption{Theory vs.\ simulations: Density evolution similar to Figure 1, but here, \( \xi \) is drawn from an exponential distribution with nonzero mean. See Example~2 for details.}
    \label{fig:density_evolution2}
\end{figure}

\paragraph{Example 2} 
In this example, we generate the vector of interest \( \boldsymbol{\xi} \) by sampling its entries independently from an exponential distribution and adding a bias of \( 0.9 \) to shift the mean. %

In Figure~\ref{fig:density_evolution2}, we compare the theoretical limiting densities with the empirical densities obtained from simulations at several time points. The results demonstrate that the theory accurately captures the mean shift and overall evolution of the density throughout the dynamics. Overall, our simulation results in Figures 1-3 confirm that the found PDE \eqref{eq:strong_pde} precisely captures the time-evaluation of the distribution of the elements of the estimates and the ODEs \eqref{eq:Q_t}--\eqref{eq:lambda_t} characterizes the time-evaluation of the state variables (cosine similarity and norm) of interest. The PDE and ODEs together capture the high-dimensional learning dynamics of the proposed algorithm \eqref{eq:dynamics_update}. In the remainder of this section, we derive practical insights (steady states, phase transitions with respect to SNR, optimal learning rate, and optimal initial norm) from the theory and then compare the algorithm with other algorithms.

\subsection{Steady-state analysis and phase transition}
In this section, we study the steady-state analysis of our algorithm using the governing coupled PDEs \eqref{eq:weak_pde} and \eqref{eq:lambda_t}. We denote the steady-state quantities as $\lambda_{s}$ and $Q_s$ in the long-time limit. To find those, we first set the right-hand sides of the \eqref{eq:Q_t} and \eqref{eq:lambda_t} to 0 and solve them together, which leads to the characterization of two cases. The first case is
\begin{equation}
    Q_{s}^2=0 \quad and \quad \lambda_{s} = \frac{1}{2} \left( 1 + \sqrt{1+2\tau} \right),
\end{equation}
which corresponds to an unstable state without learning. The other case is as follows:
\begin{equation}
    Q_{s}^2 = \frac{\omega^2 + \omega - \tau/2}{\omega^2 + \omega + \tau\omega/2}  \quad and \quad \lambda_{s} = \omega + 1,
    \label{eq:learning_capacity}
\end{equation}
which indicates the limiting performance ($Q_t$) of the INO-PCA algorithm and the norm $\lambda_{s}$ converges to the leading eigenvalue $\omega + 1$ in alignment with our earlier explanation about the algorithm. Note that the limiting performance reveals that a vanishing learning $\tau \to 0$ is required to achieve perfect estimation (i.e., $Q_s=1$), which is consistent with the behavior of other SGD-type PCA algorithms such as Oja's method. Moreover, the formula also indicates
that the algorithm is unable to learn (i.e., $Q_{s}^2=0 $) when $\omega \in \Big(0, \omega_c= \frac{-1 + \sqrt{1+2\tau}}{2}\Big)$, and a simple phase transition phenomenon occurs at $\omega_c$. 

Next, we derive the steady state density by assigning the right-hand side of \eqref{eq:strong_pde} to 0. Then, we integrate both sides so that the resulting equation is a first-order homogeneous ODE. Then, the steady state density (the solution of the ODE) is as follows:
\begin{equation}
     P_{s}(x|\xi) = \frac{1}{Z} e^{ \frac{\tau}{J(Q_s)} \left(  2\omega Q_s {\xi} x + \frac{{x^2}}{\lambda_s} - {x^2} \right)}
    \label{eq:steady_state_dist}
\end{equation}
where $Q_s$ and $\lambda_s$ are $Q$ and $\lambda$ in the steady state respectively and $Z$ is the normalization constant.

\begin{figure}[htb]
    \centering
        \begin{subfigure}[b]{0.45\textwidth}
         \centering
         \includegraphics[width=0.99\textwidth]{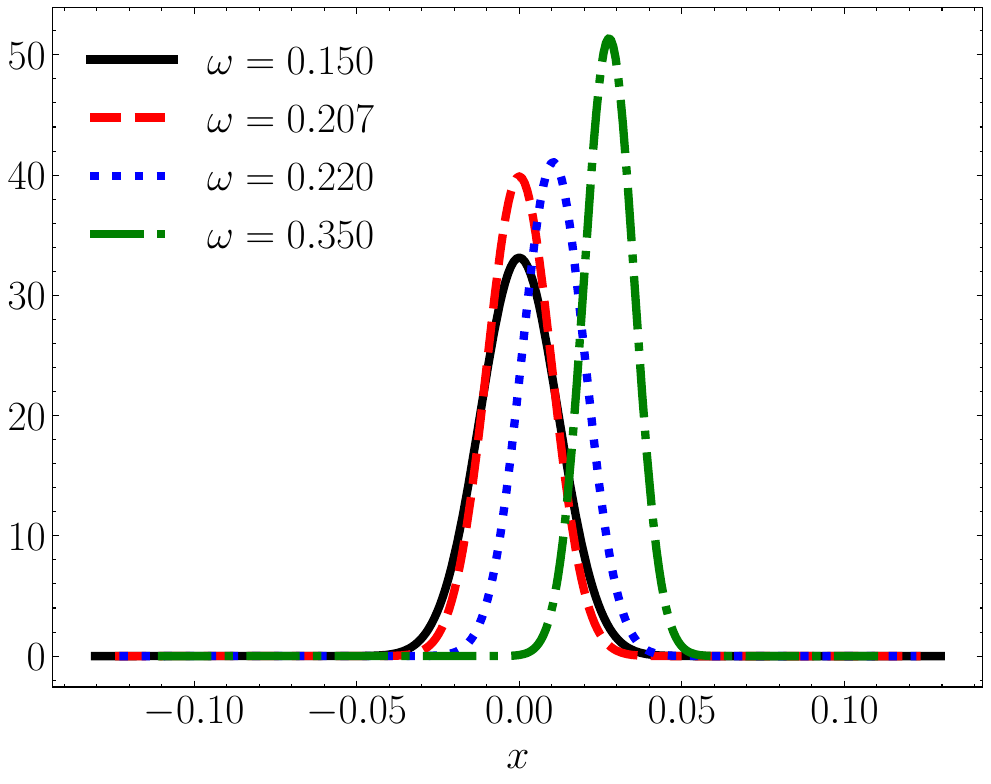}
     \end{subfigure}
     \begin{subfigure}[b]{0.45\textwidth}
         \centering
         \includegraphics[width=0.99\textwidth]{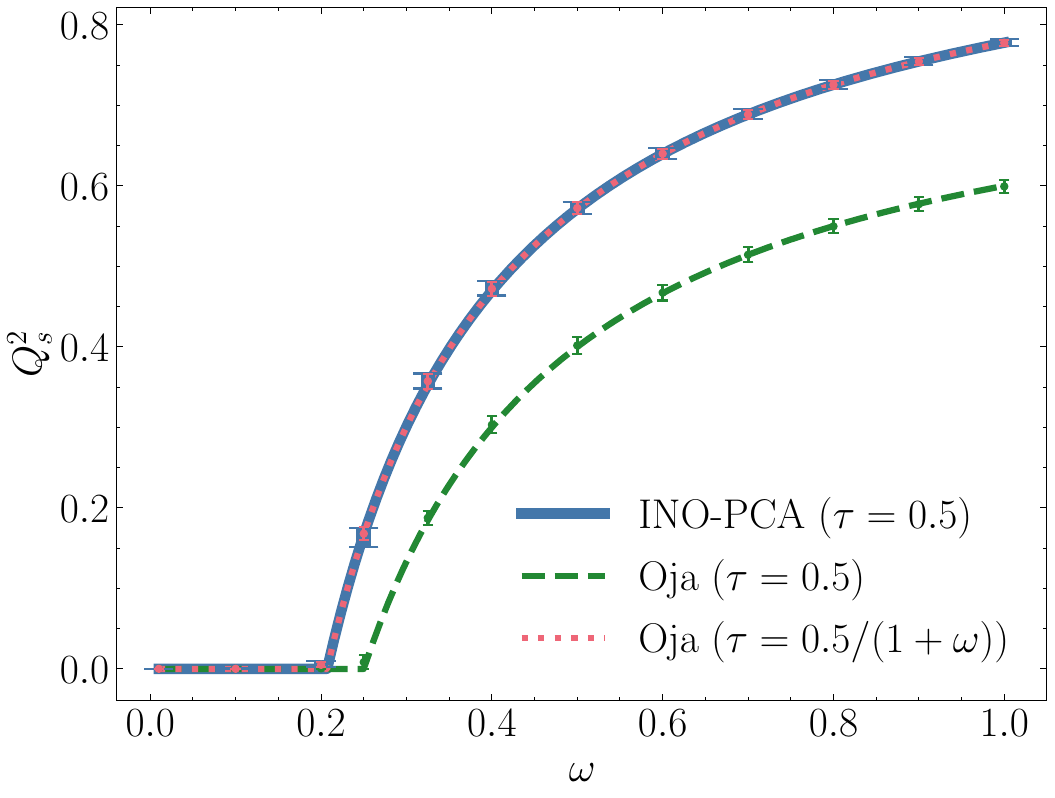}
     \end{subfigure}
    \caption{Steady-state distributions and phase transitions. Left-hand side: The
steady-state densities $P_{s}(x|\xi= 1/\sqrt{0.05})$ for different values of the SNR parameter $\omega$ (Example 3). Right-hand side: Theoretical predictions of the $Q_s$ as a function of the SNR parameter $\omega$.}\label{fig:steady_state_dist}
\end{figure}

\paragraph{Example 3} To show the the phase transition phenomenon in the steady state densities, we consider that elements of $\bxi$ is generated from a mixture distribution:
$$\xi^i \sim \pi(\xi) = (1-\rho) \delta(\xi) + \rho \delta(\xi - 1/\sqrt{\rho}) \quad \forall i \in \{1,\dots,p\},$$
where $\rho$ is the sparsity level set to 0.05 as in the sparse setting of \cite{Wang2016}. 

Figure \ref{fig:steady_state_dist} demonstrates the steady-state distributions and the phase transition phenomenon. On the left-hand side, we plot the steady state densities based on equation \eqref{eq:steady_state_dist} for the case of Example 3. We observe that the steady-state distribution starts to shift towards the true distribution after the phase transition point at $\omega_c$. On the right-hand side, we show the steady-state cosine similarity values with respect to the SNR parameter $\omega$. A clear phase transition appears at a critical value $\omega_c$. The theoretical prediction $\omega_c = 0.207$ matches well with the average of Monte Carlo simulations of the algorithm over 10 realizations (bars indicate one standard deviation). Comparing our method with Oja’s method, we see that our method has a lower phase transition threshold and achieves a higher $Q_s$. Also, we find that to achieve the same level of steady-state cosine similarity as our method when using Oja's method, one should set $\tau = 0.5/(1+\omega)$. While Oja's method can achieve the same steady-state cosine similarity using smaller learning, it reaches the steady-state later than our method, as illustrated in Figure \ref{fig:comparison} when comparing the learning curves of the two methods. Overall, this steady-state analysis reveals a relationship between the learning rate $\tau$ of our algorithm and that of  Oja's method, while we utilize this relationship to provide a fair comparison (in terms of learning rates) in Figure \ref{fig:comparison}, which demonstrates that our method (INO-PCA) is significantly faster compared to Oja's algorithm.

\begin{figure}[htb]
    \centering
    \includegraphics[width=0.5\textwidth]{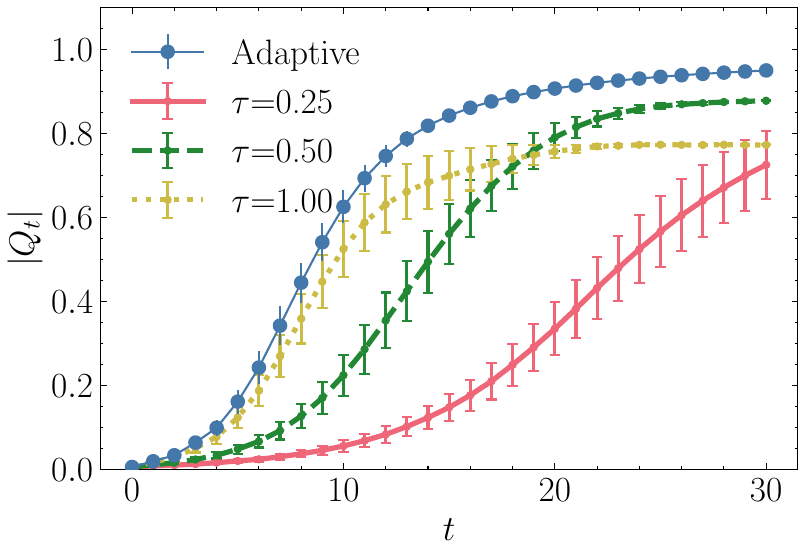}
    \caption{Comparison of the cosine similarities for INO-PCA with fixed learning rates $\tau$ and adaptive INO-PCA with learning rate given by \eqref{eq:optimal_nu_t} for $\lambda_0 = 1$ where the bars indicate one-third standard deviation.}
     \label{fig:tau}
\end{figure}

\subsection{Optimal adaptive learning rate: a three-way relationship between the norm, SNR, and learning rate}
\label{sec:optimal_steady}
In this section, we study the learning rate $\tau$ of the INO-PCA algorithm in detail with the goal of deriving further insights from our theoretical characterization. Specifically, we address how the found ODE \eqref{eq:Q_t} for cosine similarity can be used to determine optimal learning rate. 

We first note that the learning rate $\tau$ and norm parameter $\lambda$ are related. Specifically, these parameters only appear as a ratio in \eqref{eq:Q_t}. Therefore, we redefine it as $\nu_t = \tau_t/\lambda_t$ (can be considered as an effective learning) and propose to maximize the instantaneous increase in the cosine similarity $Q_t$ in terms of $\nu_t$. This approach leads to the following optimization:
\begin{equation}
    \hat{\nu_t}=\argmax_{\nu_t} \quad \nu_t Q_t (\omega - \omega Q_t^2  - \frac{\nu_t(\omega Q_t^2 + 1)}{2}),
    \label{eq:max_nu_t}
\end{equation}
which has the optimal solution given as follows: %
\begin{equation}
   \hat{\nu_t} = \frac{\tau_t}{\lambda_t} = \frac{\omega (1-Q_t^2)}{\omega Q_t^2 + 1}.
   \label{eq:optimal_nu_t}
\end{equation}
This optimal solution \eqref{eq:optimal_nu_t} demonstrates a three-way relationship between the norm parameter $\lambda_t$, SNR parameter $\omega$, and adaptive learning rate $\tau_t$, while highlighting the significant effect of the norm $\lambda_t$ on the learning dynamics. In what follows, we study the 
impact of this result first on an (oracle) adaptive learning rate $\tau_t$ here and then on an optimal initialization of the norm $\lambda_0$ in the next subsection. 

\begin{rem}
   The adaptive rule derived here depends on the instantaneous values of $Q_t$ and the signal strength $\omega$, which are not directly observable. As such, this rule should be interpreted as an "oracle benchmark" rather than a practical algorithm. Its purpose is to elucidate the role of norm-dependent scaling in regulating the effective learning rate and to characterize the best achievable dynamics within this class of updates. 
\end{rem}

Suppose the time-evolving $\hat{\nu_t}$ in \eqref{eq:optimal_nu_t} is achieved by an adaptive (time-varying) learning rate $\tau_t$, which leads to an adaptive (in terms of the learning rate) version of the algorithm, and we call it "adaptive INO-PCA". In Figure \ref{fig:tau}, we numerically show the optimality of adaptive $\tau_t$ compared against various fixed $\tau$ values for $\lambda_0=1$. Clearly, the method with the adaptive learning rate outperforms the fixed learning rate cases.

\begin{figure}[htb]
    \centering
        \begin{subfigure}[b]{0.55\textwidth}
         \centering
         \includegraphics[width=0.99\textwidth]{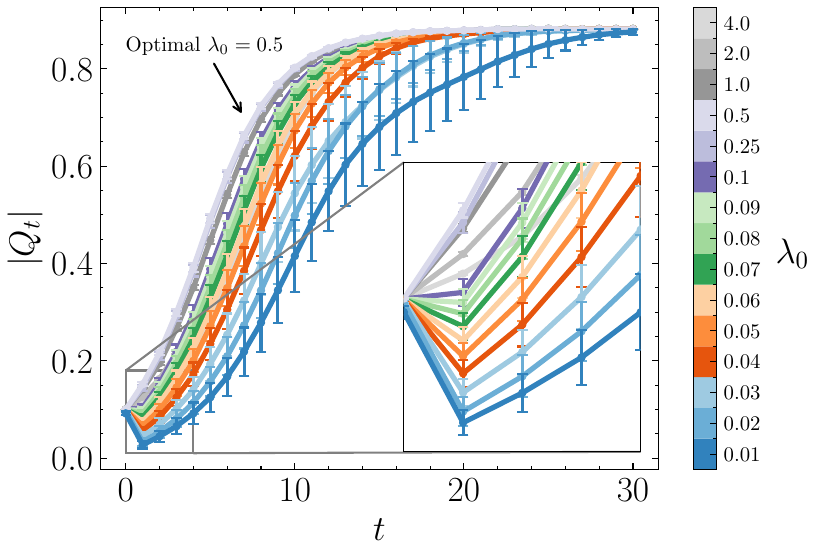}
     \end{subfigure}
     \begin{subfigure}[b]{0.39\textwidth}
         \centering
         \includegraphics[width=0.99\textwidth]{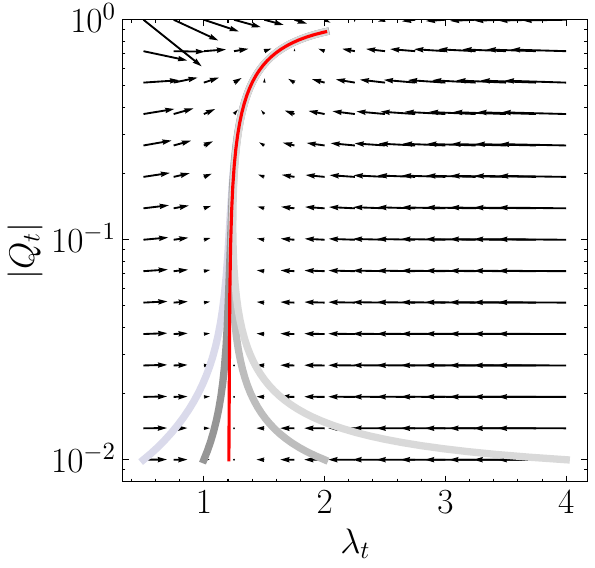}
     \end{subfigure}
    \caption{Comparison of the cosine similarities for different $\lambda_0$ values with $\tau = 0.5$ and $t=30$. On the left, numerical simulations are plotted, where the bars indicate one-third standard deviation for the Monte Carlo simulations. On the right, so-called "phase portrait" of $Q_t$ and $\lambda_t$ is illustrated based on \eqref{eq:Q_t} and \eqref{eq:lambda_t}. The curves in the portrait represent trajectories under different $\lambda_0$ initializations, with $Q_0 = 10^{-2}$. The red curve is for $\lambda_0=\frac{1}{2}(1+\sqrt{1+2\tau})$.}
    \label{fig:lambda0}
\end{figure}

\subsection{The choice of the initial norm of the estimate ($\lambda_0$)}
Next, we examine the role of the initialization scale \( \lambda_0 \) in greater detail. Figure~\ref{fig:lambda0} illustrates how different choices of \( \lambda_0 \) affect the evolution of the cosine similarity. First, observe that the steady-state cosine similarity \( Q_s \) is independent of the initial norm \( \lambda_0 \). This can be seen from the phase portrait of \( (Q_t, \lambda_t) \), plotted using the ODEs \eqref{eq:Q_t}--\eqref{eq:lambda_t} (right panel of Figure~\ref{fig:lambda0}), where trajectories initialized at different \( \lambda_0 \) values are all attracted toward the trajectory emanating from \( \lambda_0 = \tfrac{1}{2}(1 + \sqrt{1 + 2\tau}) \). The convergence and eventual merging of these trajectories confirm that the steady-state cosine similarity does not depend on \( \lambda_0 \).

On the other hand, the value of \( \lambda_0 \) has a clear effect on the early-stage learning speed, as shown in the left panel of Figure~\ref{fig:lambda0}. In particular, we observe the existence of an "optimal" initialization scale that yields the fastest initial increase in cosine similarity. Using the same analysis employed in deriving the optimal learning rate in \eqref{eq:max_nu_t}--\eqref{eq:optimal_nu_t}, together with the fact that the initial cosine similarity is close to zero, we find that the optimal choice is \( \lambda_0 = \tau/\omega \). The numerical results in Figure~\ref{fig:lambda0} (left) confirm this prediction: for the parameters used in this experiment, the optimal initialization occurs near \( \tau/\omega = 0.5 \).

We additionally note that the left panel of Figure~\ref{fig:lambda0} is shown for the warm-start setting \( Q_0 = 0.1 \), which helps reveal how poorly chosen initialization of the norm can lead to temporary degradation in performance (a drop in cosine similarity) before the algorithm eventually recovers and resumes its typical learning trajectory.

\subsection{Comparison of algorithms}
\label{sec:comparison}
In this section, we first compare the performance of our algorithm (INO-PCA) with that of Oja’s method. We then evaluate the adaptive variant of our algorithm (Adaptive INO-PCA) against representative baselines that are Candid Covariance-Free Incremental PCA (CCIPCA) algorithm \citep{Weng2003,Zhao2006} and AdaOja \citep{henriksen2019adaoja} in a non-stationary environment, where the vector of interest \( \boldsymbol{\xi} \) changes midway through the learning process. For each method, we report the time evolution of the cosine similarity \( |Q_t| \), which provides a direct and interpretable measure of tracking accuracy over time.

\begin{rem}
    Our empirical evaluation focuses on comparisons with Oja’s algorithm to isolate and highlight the effects of implicit normalization and norm dynamics. A comprehensive empirical comparison with other online PCA methods is therefore beyond the scope of the present analytical study, and we view it as an interesting direction for future work.
\end{rem}

\begin{figure}[htb]
    \centering
     \includegraphics[width=0.5\textwidth]{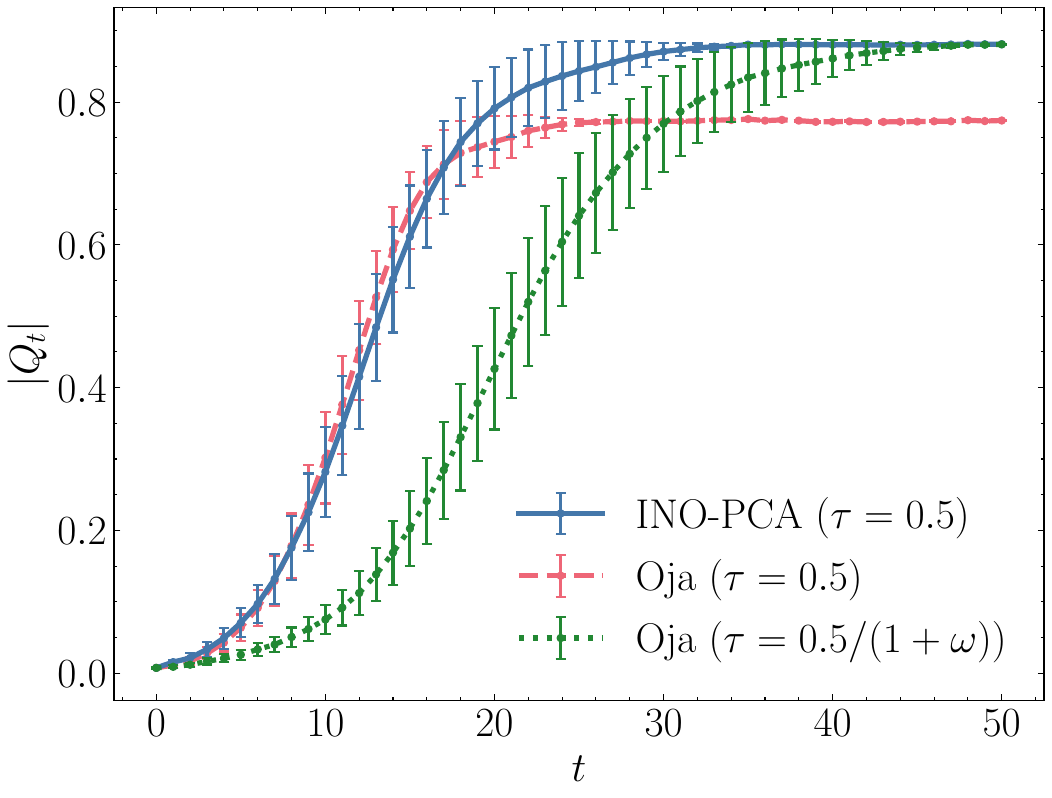}
    \caption{A comparison of the learning ($Q_t$) curves (with bars indicating one-third standard deviation) of the algorithms for the Monte Carlo simulations. $\lambda_0$ for INO-PCA is set to $0.5$.}
    \label{fig:comparison}
\end{figure}

\subsubsection{INO-PCA vs. Oja's Algorithm}
For the setting in Example~1 described above, we compare INO-PCA with the classical stochastic algorithm of Oja. Our focus here is on SGD-type online PCA methods, since INO-PCA also belongs to this class. Oja's algorithm is a canonical representative of this family, and while one could also include methods such as Krasulina's algorithm or other SGD variants, Oja's method is known to perform equivalently to Krasulina's under the present conditions \citep{Balsubramani2013}. Moreover, INO-PCA is structurally closely related to Oja's update, making a direct comparison both natural and informative. For these reasons, we restrict attention to Oja's method for clarity of exposition.

Figure~\ref{fig:comparison} reports the experimental results. The plot shows that INO-PCA behaves similarly to Oja's algorithm with learning rate \( \tau = 0.5 \) during the early phase of learning. However, its steady-state cosine similarity \( Q_s \) matches that of Oja's algorithm with learning rate \( \tau = 0.5/(1+\omega) \). In this sense, INO-PCA combines the benefits of both regimes: it learns as quickly as Oja's algorithm with a relatively large step size in the initial iterations, while ultimately achieving the same steady-state performance as Oja's method with a more conservative learning rate. Overall, this comparison demonstrates that INO-PCA converges faster than Oja's algorithm while implicitly adapting to the effective learning-rate scaling induced by its update rule.

\begin{figure}[htb]
    \centering
     \includegraphics[width=0.9\textwidth]{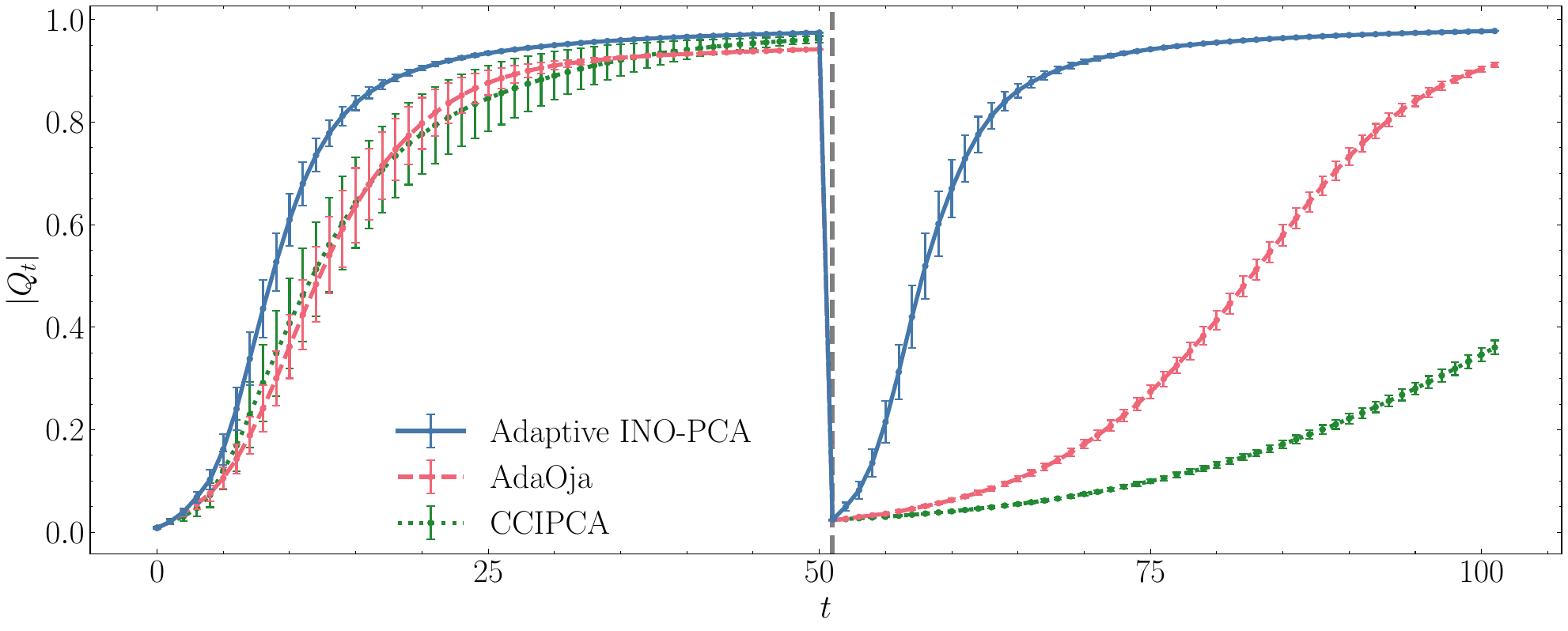}
    \caption{Adaptation behavior when the vector of interest \( \boldsymbol{\xi} \) is changed abruptly at \( t = 50 \). The initial \( \boldsymbol{\xi} \) is sampled from the distribution used in Example~1 (non-sparse), while the second \( \boldsymbol{\xi} \) is sampled from the sparse version of the same example. The learning curves \( Q_t \) (with error bars indicating one-third standard deviation) are shown for each method. The amnesic parameter of CCIPCA is empirically selected and set to \( 4 \).}
    \label{fig:adaptive_comparison}
\end{figure}

\subsubsection{Adaptive INO-PCA vs. representative baselines}
Finally, we evaluate the performance of our adaptive algorithm (Adaptive INO-PCA) in comparison with CCIPCA \citep{Weng2003} and AdaOja \citep{henriksen2019adaoja}. AdaOja augments Oja’s method with an adaptive learning rate, while CCIPCA estimates principal components via iterative averaging. We include CCIPCA due to its structural similarity to our approach, as discussed in Section~\ref{sec:problem-and-algorithm}. To assess the adaptability of these methods, we consider a setting in which the vector of interest \( \boldsymbol{\xi} \) changes abruptly during learning, allowing us to observe how quickly each algorithm responds to a non-stationary environment. Such scenarios arise naturally in applications like online sensing or autonomous driving, where data distributions may shift suddenly due to external factors such as weather or illumination.

We use the same distribution as in Example~1 to generate the initial \( \boldsymbol{\xi} \), and the distribution in Example~3 to generate the second \( \boldsymbol{\xi} \). For \( t \in [0,50] \), the first \( \boldsymbol{\xi} \) is used; for \( t \in [51,100] \), the second \( \boldsymbol{\xi} \) is used.

Figure~\ref{fig:adaptive_comparison} presents the results for this setting. During the first phase \( t \in [0,50] \), Adaptive INO-PCA and the baseline methods achieve comparable initial and steady-state cosine similarity, although Adaptive INO-PCA exhibits noticeably faster improvement in the middle of the interval, reflecting the advantage of its adaptive learning rate. During the second phase \( t \in [51,100] \), Adaptive INO-PCA rapidly regains high cosine similarity after the change in \( \boldsymbol{\xi} \), whereas AdaOja and CCIPCA adapt much more slowly. Because both AdaOja and CCIPCA progressively reduce the influence of new samples over time, their updates become sluggish in the face of sudden distribution shifts. In contrast, Adaptive INO-PCA maintains a high degree of responsiveness, enabling it to track the new principal component effectively.

\subsubsection{Extension to real-world data}
\begin{figure}[htb]
    \centering
        \begin{subfigure}[b]{0.48\textwidth}
         \centering
         \includegraphics[width=0.99\textwidth]{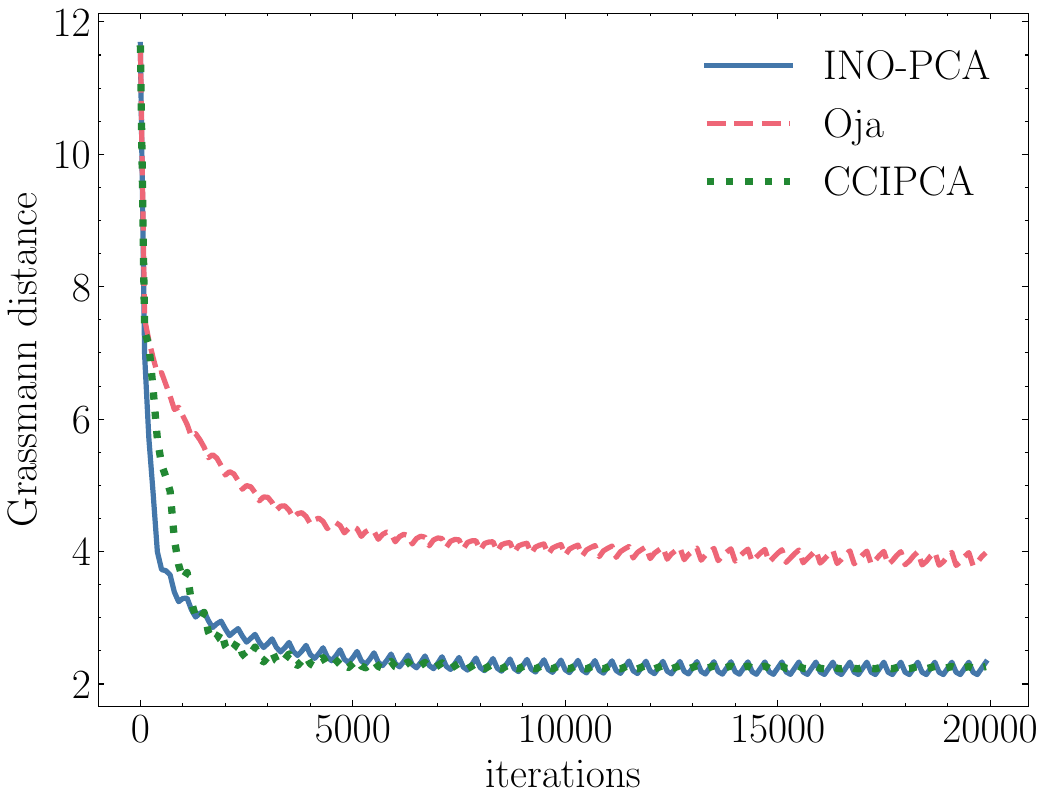}
     \end{subfigure}
     \hfill
     \begin{subfigure}[b]{0.45\textwidth}
         \centering
         \includegraphics[width=0.99\textwidth]{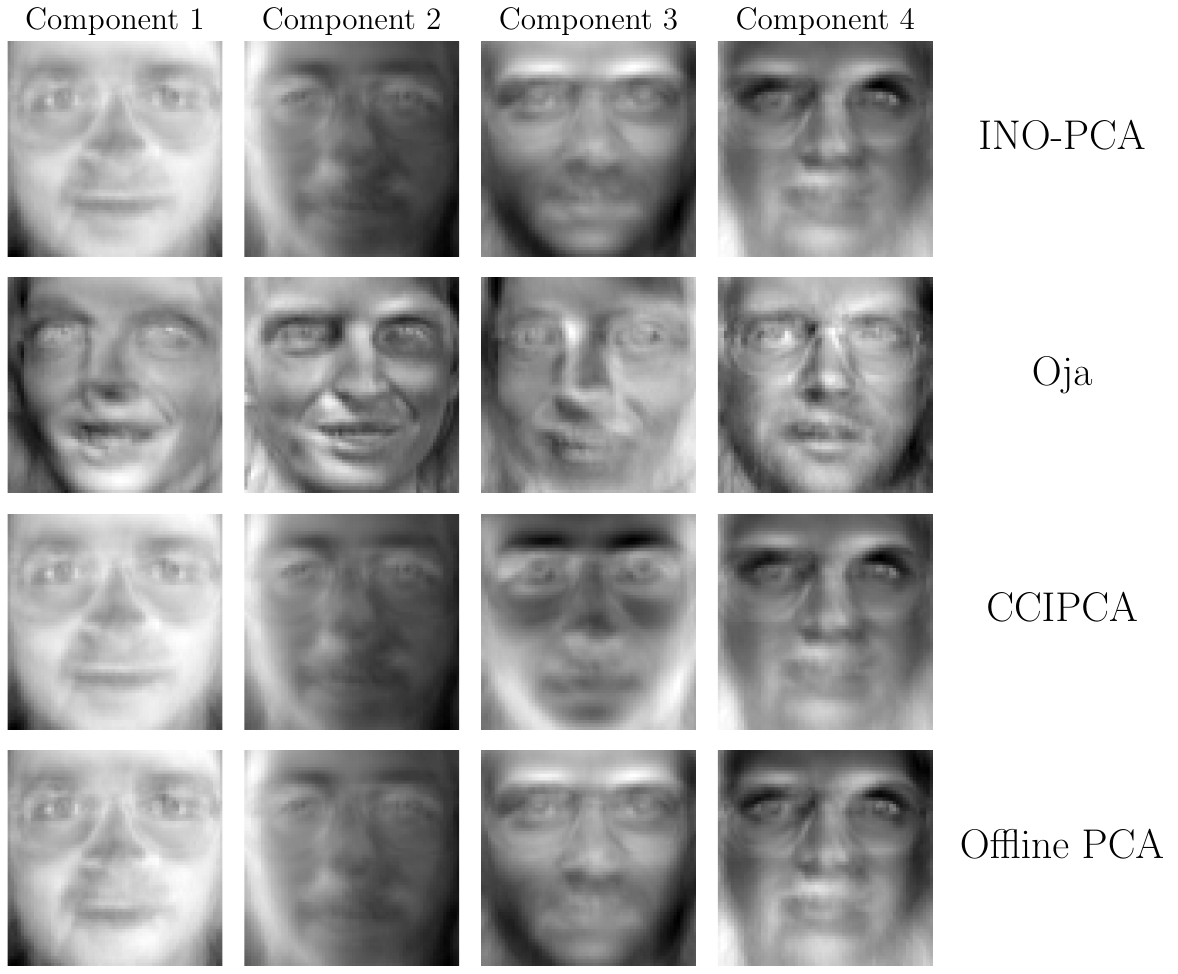}
     \end{subfigure}
    \caption{Comparison on a real-world subspace learning task using the Olivetti Faces dataset for INO-PCA, Oja's method, and CCIPCA. Left: time evolution of the Grassmann distance between the estimated subspace and the true subspace, where the true subspace is approximated using the offline PCA implementation in scikit-learn \citep{scikit-learn}. Right: estimates of the first principal components obtained by each method. The learning rates for Oja's method and INO-PCA, as well as the amnesic parameter for CCIPCA, are selected via grid search with the objective of minimizing the Grassmann distance at the 4000-th iteration. }\label{fig:olivetti}
\end{figure}

To demonstrate the practical usefulness of the proposed method (INO-PCA) on a real-world task, we compare its performance with Oja's algorithm and CCIPCA on a subspace learning problem using the Olivetti Faces dataset \citep{ATT_FaceDatabase}, as shown in Figure~\ref{fig:olivetti}. Following prior work in subspace learning \citep{bond2024exploring}, we measure the discrepancy between the true and estimated subspaces using the Grassmann distance. Multiple principal components are estimated using the extension of INO-PCA to the multi-component setting described in Appendix~\ref{appendix:extention_to_multiplePC}. 

Figure~\ref{fig:olivetti} (left) shows that INO-PCA learns the principal components quickly and accurately with respect to the Grassmann distance. The corresponding estimated components are visualized in Figure~\ref{fig:olivetti} (right). These results indicate that INO-PCA can outperform CCIPCA in the early stages of learning while ultimately achieving comparable steady-state performance. At the same time, INO-PCA substantially outperforms Oja's method throughout. Overall, the experiment confirms that INO-PCA is practically effective, particularly when only a small number of iterations are available or when robustness to abrupt changes is required.

\section{Informal derivation of the main theoretical results}
\label{sec:exchangebility_and_derivation}

We now provide an informal derivation of our main theoretical results for the INO-PCA algorithm. We begin by introducing the notion of exchangeability \citep{Diaconis1977,Diaconis1980,Aldous1985}, which is a key structural property underlying our analysis, and then outline the derivation of the PDE \eqref{eq:weak_pde}. A complete and rigorous proof of Theorem~\ref{theo:main_conv} is given in Appendices~\ref{appendix:meta-theorem} and~\ref{appendix:formal_proof}.

Let \( M_k = [\boldsymbol{x}_k, \boldsymbol{\xi}] \) denote the Markov state of the algorithm update \eqref{eq:dynamics_update}. We first observe that \( M_k \) forms an exchangeable Markov chain on \( (\mathbb{R}^2)^{\otimes p} \) governed by the update equation \eqref{eq:dynamics_update}. Exchangeability plays a central role in the analysis of high-dimensional stochastic systems: it implies that the joint evolution of the coordinates is invariant under permutations, and therefore the large-scale behavior of the process can be characterized through the evolution of its empirical measure. This property enables us to decouple the coordinate-wise dynamics in the asymptotic limit, following the mean-field approach developed in prior work \citep{wang2017scaling}.

\begin{definition}
A joint distribution \( P(\boldsymbol{x}) \) is said to be \emph{exchangeable} if  
\[
P(\mathcal{P}_\pi \boldsymbol{x}) = P(\boldsymbol{x})
\]
for any permutation matrix \( \mathcal{P}_\pi \) and any vector \( \boldsymbol{x} = [x^1, x^2, \dots]^T \). In other words, the distribution is invariant under arbitrary coordinate permutations.
\end{definition}

In our setting, we require an extension of this notion to Markov chains.

\begin{definition}
Let \( \mathbb{S} \) be a Polish space (for example, \( \mathbb{R}^2 \)). For any permutation matrix \( \mathcal{P}_\pi \), any Borel set \( \mathcal{B} \subset \mathbb{S}^{\otimes p} \), and any state \( \boldsymbol{m} \in \mathbb{S}^{\otimes p} \), a Markov chain \( \{\boldsymbol{m}_k\} \) is \emph{exchangeable} if
\begin{equation}
    P(\boldsymbol{m}_{k+1} \in \mathcal{B}_\pi \mid \boldsymbol{m}_k = \mathcal{P}_\pi \boldsymbol{m})
    =
    P(\boldsymbol{m}_{k+1} \in \mathcal{B} \mid \boldsymbol{m}_k = \boldsymbol{m}),
\end{equation}
where \( \mathcal{B}_\pi = \{ \mathcal{P}_\pi \boldsymbol{m} : \boldsymbol{m} \in \mathcal{B} \} \). This condition ensures that permuting the coordinates of the state results in an equivalent permutation of the transition behavior.
\end{definition}

Since \( \boldsymbol{\xi} \) is fixed and the initialization \( \mathbf{x}_0 \) is assumed to be exchangeable, it suffices to verify that the update rule preserves exchangeability for \( \mathbf{x}_k \). Observe that  
\( \mathbf{m}_{k+1} \in \mathcal{B}_\pi \iff \mathcal{P}_\pi^T \mathbf{m}_{k+1} \in \mathcal{B} \).  
Using the derivation in Appendix~\ref{section:detailed_exchangability}, we obtain for any permutation matrix \( \mathcal{P}_\pi \)
\begin{align}
    \mathcal{P}_\pi^T \pmb{{x}}_{k+1} = \pmb{{x}}_k + \frac{\tau}{p} \left(\frac{ \left( \sqrt{\frac{\omega}{p}} c_k \pmb{\xi} + \mathcal{P}_\pi^T \mathbf{a}_k \right) \pmb{{x}}_k^T \left( \sqrt{\frac{\omega}{p}} c_k \pmb{\xi} + \mathcal{P}_\pi^T \mathbf{a}_k \right)}{\lambda_k} - \pmb{{x}}_k \right).
    \label{eq:exchangeability}
\end{align}
Since \( \boldsymbol{a}_k \sim \mathcal{N}(0, \mathbf{I}) \) is an exchangeable random vector and the update \eqref{eq:exchangeability} is invariant under permutations, the Markov chain \( \{ \mathbf{m}_k \} \) is exchangeable. For additional background on exchangeability, see \citet{wang2017scaling, Diaconis1977, Diaconis1980, Aldous1985}.

Next, we derive the PDE \eqref{eq:weak_pde}. Let
\[
\boldsymbol{\Delta}_k
= \boldsymbol{x}_{k+1} - \boldsymbol{x}_k
= \frac{\tau}{p} \left( \mathbf{y}_{k}\mathbf{y}_{k}^T \frac{\boldsymbol{x}_k}{\lambda_k} - \boldsymbol{x}_k \right).
\]
We first compute the first- and second-order conditional moments of \( \boldsymbol{\Delta}_k \). In the literature (for example, \citet{Wang2016}), these are referred to as the drift and diffusion terms, respectively. All expectations in the following derivations are conditional on the sigma-field \( \mathcal{F}_k^p \) generated by \( \{\boldsymbol{\xi}, \boldsymbol{x}_0, \dots, \boldsymbol{x}_k\} \), and we use the shorthand
\( \mathbb{E}_k[\cdot] = \mathbb{E}[\cdot \mid \mathcal{F}_k^p] \). After a detailed calculation, provided in Appendix~\ref{sec:derive_drift_diffusion}, we obtain
\begin{align}
    \mathbb{E}_k\left[\mathbf{\Delta}_{k}\right] &= \frac{\tau}{p} \left(\omega Q_k \pmb{\xi} +  \frac{\pmb{x}_k}{\lambda_k} - \pmb{x}_k\right),    \label{eq:drift}\\
    \mathbb{E} \left[ \mathbf{\Delta}_{k} \mathbf{\Delta}_{k}^T \right] &= \frac{\tau^2}{p}(\omega Q_k^2 + 1) I + \mathcal{O}(1/p^2).
    \label{eq:diffusion}
\end{align}
In both the first-order moment \eqref{eq:drift} and the second-order moment \eqref{eq:diffusion}, the leading terms are of order \( 1/p \). This indicates that the characteristic time scale of the Markov process is of order \( 1/p \). Consequently, the higher-order terms in \eqref{eq:diffusion} can be neglected in the scaling limit, and the continuous-time embedding in \eqref{eq:continous_embedding} is naturally chosen with the time-rescaling \( k = \lfloor p t \rfloor \).

Using the conditional moments derived above, we now obtain the PDE \eqref{eq:weak_pde}. Let \( f(x,\xi) \) be a test function satisfying the stated regularity and boundedness conditions. Applying a Taylor expansion in the \( x \)-coordinate, we have
\begin{align}
    f\left(x_{k+1}^{i}, \xi\right) &= f\left(x_{k}^{i}+\Delta_{k}^{i}, \xi \right) \nonumber \\
    &= f\left(x_{k}^{i}, \xi \right)+\frac{\partial f}{\partial x}\left(x_{k}^{i}, \xi \right) \Delta_{k}^{i} + \frac{1}{2!} \frac{\partial^2 f}{\partial x^2}\left(x_{k}^{i}, \xi \right) (\Delta_{k}^{i})^2 + h_k^i,
\end{align}
where
\[
    h_k^i
    = \frac{1}{3!} \frac{\partial^3 f}{\partial x^3}\big(c_{k}^{i}, \xi \big)\, (\Delta_{k}^{i})^3
\]
for some \( c_{k}^{i} \in [x_{k}^{i}, x_{k}^{i} + \Delta_{k}^{i}] \) is the higher-order remainder term in Lagrange form.

We now express the functional inner product in terms of the empirical measure:
\begin{align}
    \left\langle f, \mu_{k+1}\right\rangle &= \frac{1}{p} \sum_{i} f(x_{k+1}^{i}, \xi)\\
    &= \left\langle f, \mu_{k} \right\rangle + \frac{1}{p} \sum_{i=1}^{p} \frac{\partial f}{\partial x}\left(x_{k}^{i}, \xi \right) \Delta_{k}^{i} + \frac{1}{p} \sum_{i=1}^{p} \frac{1}{2!} \frac{\partial^2 f}{\partial x^2}\left(x_{k}^{i}, \xi \right) (\Delta_{k}^{i})^2 + \frac{1}{p} \sum_{i=1}^{p} h_k^i \nonumber\\
    &= \left\langle f, \mu_{k} \right\rangle +  \frac{1}{p} \sum_{i=1}^{p} \mathbb{E} \left[\frac{\partial f}{\partial x}\left(x_{k}^{i}, \xi \right) \Delta_{k}^{i} \right] + \frac{1}{p} \sum_{i=1}^{p} \frac{1}{2!} \mathbb{E} \left[\frac{\partial^2 f}{\partial x^2}\left(x_{k}^{i}, \xi \right) (\Delta_{k}^{i})^2 \right] + \bar{h}_k + \bar{m}_k,\nonumber
\end{align}
where we define
\begin{align}
   \bar{h}_k \stackrel{\text{def}}{=} \frac{1}{p}\sum_{i=1}^{p} h_k^i, \quad \textit{and} \quad \bar{m}_k \stackrel{\text{def}}{=} &\Bigg\{\frac{1}{p}\sum_{i=1}^{p} \frac{\partial f}{\partial x}\left(x_{k}^{i}, \xi \right) \Delta_{k}^{i} + \frac{1}{p}\sum_{i=1}^{p} \frac{1}{2!} \frac{\partial^2 f}{\partial x^2}\left(x_{k}^{i}, \xi \right) (\Delta_{k}^{i})^2 \nonumber\\
   &- \frac{1}{p}\sum_{i=1}^{p} \mathbb{E} \left[\frac{\partial f}{\partial x}\left(x_{k}^{i}, \xi \right) \Delta_{k}^{i} \right] + \frac{1}{p}\sum_{i=1}^{p} \frac{1}{2!} \mathbb{E} \left[\frac{\partial^2 f}{\partial x^2}\left(x_{k}^{i}, \xi \right) (\Delta_{k}^{i})^2 \right] \Bigg\}. \nonumber
\end{align}
Here \( \bar{h}_k \) collects the higher-order Taylor terms, while \( \bar{m}_k \) captures the martingale fluctuations around the conditional expectations. These terms typically vanish in the scaling limit $p \to \infty$. In the last step, we introduced conditional expectation terms and defined \( \bar{m}_k \) as the deviation from these expectations, since the latter can be computed explicitly using \eqref{eq:drift} and \eqref{eq:diffusion}.

\noindent Using the law of total expectation together with the moment expressions \eqref{eq:drift}--\eqref{eq:diffusion}, we obtain
\begin{align}
    \left\langle f, \mu_{k+1}\right\rangle &- \left\langle f, \mu_{k} \right\rangle = \frac{1}{p} \left\langle G({x}, \lambda, {\xi}, Q) \frac{\partial f}{\partial x}, \mu_k \right\rangle + \frac{1}{2p} \left\langle J(Q) \frac{\partial^2 f}{\partial x^2}, \mu_k \right\rangle + \mathcal{O}(1/p^2) + \bar{h}_k + \bar{m}_k, \nonumber
\end{align}
where
\begin{equation}
    G({x}, \lambda, {\xi}, Q) \stackrel{\text{def}}{=} \tau(\omega Q {\xi} +  \frac{{x}}{\lambda} - {x}),\quad
    J(Q) \stackrel{\text{def}}{=} \tau^2(\omega Q^2 + 1).\nonumber
\end{equation}
A detailed derivation of this one-step deviation formula is provided in Appendix~\ref{sec:derivation_of_one_step_deviation}.

\noindent For convenience, define
\[
    \bar{v}_k \stackrel{\text{def}}{=}
    \big(\left\langle f, \mu_{k+1}\right\rangle - \left\langle f, \mu_{k} \right\rangle\big)
    - \bar{h}_k - \bar{m}_k.
\]
Then, we can write
\begin{align}
    \left\langle f, \mu_{k}\right\rangle - \left\langle f, \mu_{0} \right\rangle
    &= V_k + H_k + M_k,
\end{align}
where
\begin{equation}
    V_k \stackrel{\text{def}}{=} \sum_{l=0}^{k-1} \bar{v}_l,
    \qquad
    H_k \stackrel{\text{def}}{=} \sum_{l=0}^{k-1} \bar{h}_l,
    \qquad
    M_k \stackrel{\text{def}}{=} \sum_{l=0}^{k-1} \bar{m}_l.
\end{equation}
For completeness, we set \( V_0 = H_0 = M_0 = 0 \). By construction, \( \{M_k\}_{k \ge 0} \) is a martingale starting at zero, capturing the martingale fluctuations around the deterministic drift \( V_k \) and the higher-order remainder \( H_k \).

\noindent Next, we apply the continuous-time embedding with a time acceleration by a factor of \( p \), setting \( k = \lfloor pt \rfloor \):
 \begin{equation}
     \mu_t \stackrel{\text{def}}{=} \mu_{\floor{pt}}, \quad V_t \stackrel{\text{def}}{=} V_{\floor{pt}}, \quad H_t \stackrel{\text{def}}{=} H_{\floor{pt}}, \quad M_t \stackrel{\text{def}}{=} M_{\floor{pt}}. 
 \end{equation}
Each of these is a piecewise-constant càdlàg function with jumps at times of length \( 1/p \). Under this embedding, we can write
\begin{align}
    \bar{v}_k
    &= \int_{\frac{k}{p}}^{\frac{k+1}{p}} L(\mu_{\hat{t}}) d\hat{t} + \mathcal{O}(1/p^2),
\end{align}
where
\begin{equation}
    L(\mu_{\hat{t}}) \stackrel{\text{def}}{=} \left\langle G({x}, \lambda, {\xi}, Q) \frac{\partial f}{\partial x}, \mu_{\hat{t}} \right\rangle + \frac{1}{2} \left\langle J(Q) \frac{\partial^2 f}{\partial x^2}, \mu_{\hat{t}} \right\rangle.
\end{equation}
The time-scaling effectively removes the \( 1/p \) factor appearing in the drift and diffusion terms, producing a finite contribution in the limit.

\noindent Putting the pieces together yields
\begin{align}
    \left\langle f, \mu_{t}\right\rangle - \left\langle f, \mu_{0} \right\rangle
    &= V_t + H_t + M_t = \int_{0}^{t} L(\mu_{\hat{t}}) d\hat{t}  + H_t + M_t + \mathcal{O}(1/p).
\end{align} 
If \( \lim_{p\to\infty} H_t = \lim_{p\to\infty} M_t = 0 \), we obtain the limiting PDE \eqref{eq:weak_pde}. This completes the informal derivation of \eqref{eq:weak_pde}; a fully rigorous proof is provided in Appendix~\ref{appendix:formal_proof}.

\vspace{-1em}
\section{Conclusion}
We introduced INO-PCA, an online PCA algorithm that removes the unit-norm constraint and instead exploits a dynamically evolving norm as an informative internal state. Our high-dimensional analysis provides an exact PDE characterization of its dynamics, yielding closed-form ODEs that expose a tight coupling between the norm, the cosine similarity, the signal-to-noise ratio, and the optimal learning rate. This perspective reveals a sharp phase transition in steady-state recovery and clarifies the algorithmic benefits of implicit normalization. Empirically, INO-PCA consistently outperforms Oja’s method with comparable computational cost, and its adaptive variant exhibits superior tracking behavior under non-stationary conditions. Overall, our results show that allowing the norm to evolve is a principled and effective mechanism for improving stability, speed, and adaptability in online PCA and potentially in other high-dimensional streaming problems.

\acks
We acknowledge that this work was initially supported by the TÜBİTAK 2232 International Fellowship for Outstanding Researchers (No.~118C337), and later by TÜBİTAK under project 124E063 within the ARDEB 1001 program, as well as by an AI Fellowship provided by the Koç University \& İş Bank Artificial Intelligence (KUIS AI) Research Center. S.D.\ is supported by an AI Fellowship from the KUIS AI Center and a PhD Scholarship (BİDEB 2211) from TÜBİTAK.

\appendix

\section{Boundedness of $\lambda_k$}
\label{appendix:bounded_lambda_k}
Here, we show the boundedness of $\lambda_k$. First, we show that $\lambda_k \neq 0$. Then, we prove that $\lambda_k$ does not diverge.

\begin{lem}
\label{lem:lambda>0}
\begin{equation}
    \min_{k\le pT} \lambda_k > 0 
\end{equation}

\begin{proof}
$\lambda_k \ge 0$ due to the properties of the norm. We can further prove $\lambda_k \neq 0$ by contradiction using the update rule \eqref{eq:dynamics_update} as follows:

Assume $\lambda_k = 0$, then $\norm{\bx_k} = 0$ and $\bx_k = 0$. Also, let $\hat{\lambda}_k$ be any eigenvalue of $\by_k \by_k^T$. 
\begin{align}
        \bx_{k} &= \bx_{k-1} + \frac{\tau}{p} \left( \by_{k-1}\by_{k-1}^T \frac{\bx_{k-1}}{\lambda_{k-1}} - \bx_{k-1} \right) = 0\\
        &= \left(\frac{\tau}{p\lambda_{k-1}} \by_{k-1}\by_{k-1}^T - \left( \frac{\tau}{p} - 1 \right) I \right) \bx_{k-1} = 0\\
        &\implies \frac{\tau}{p} \frac{\hat{\lambda}_{k-1}}{\lambda_{k-1}} - \left(\frac{\tau}{p} - 1 \right) = 0 \quad 	\lor \quad \bx_{k-1} = 0\\
        &\implies  \frac{\hat{\lambda}_{k-1}}{\lambda_{k-1}} = 1 - \frac{p}{\tau} < 0 \quad \lor \quad \bx_{k-1} = 0 \quad \textit{(since $p>>\tau$)}
\end{align}

Assuming $\bx_{\hat{k}} \neq 0$ for $\hat{k} \in \{0,1, \dots, k-1\}$, we get $\frac{\hat{\lambda}_{k-1}}{\lambda_{k-1}} < 0$.

\begin{align}
    &\lambda_{k-1} \ge 0 \quad \textit{(since $\lambda_k = \norm{\bx_k}/\sqrt{p}$)}\\
    &\implies \hat{\lambda}_{k-1} < 0
    \label{eq:y_k_eigenvalue}
\end{align}

Since $\by_k \by_k^T$ is positive semi-definite ($z^T \by_k \by_k^T z = (\by_k^T z)^2 \ge 0$ for all $z \in \mathcal{R}^p - \{0\}$) and symmetric, all of its eigenvalues are non-negative. This contradicts with \eqref{eq:y_k_eigenvalue}. Therefore, $\lambda_k \neq 0$. 
\end{proof}
\end{lem}

\begin{lem}
\label{lem:lambda_bound}
\begin{equation}
    C_1(T) \le \max_{k \le pT} \lambda_k \le C_2(T)
\end{equation}

\begin{proof}
Let $\{\hat{\lambda}_k^1, \hat{\lambda}_k^2, \dots, \hat{\lambda}_k^p\}$ (with $\hat{\lambda}_k^1 \ge \hat{\lambda}_k^2 \ge \dots \ge \hat{\lambda}_k^p \ge 0$) be the set of eigenvalues of $\by_k \by_k^T$. Here, $\hat{\lambda}_{k-1}^i \ge 0$ because $\by_k \by_k^T$ is positive semi-definite ($z^T \by_k \by_k^T z = (\by_k^T z)^2 \ge 0$ for all $z \in \mathcal{R}^p - \{0\}$) and symmetric. Then, we show that the update rule is a linear mapping as:
\begin{align}
        \bx_{k} &= \bx_{k-1} + \frac{\tau}{p} \left( \by_{k-1}\by_{k-1}^T \frac{\bx_{k-1}}{\lambda_{k-1}} - \bx_{k-1} \right)\\
        &= \left(\frac{\tau}{p\lambda_{k-1}} \by_{k-1}\by_{k-1}^T - \left( \frac{\tau}{p} - 1 \right) I \right) \bx_{k-1} = M_{k} \bx_{k-1}\\
        &\implies \Lambda_{k}^i = \frac{\tau}{p} \frac{\hat{\lambda}_{k-1}^i}{\lambda_{k-1}} - \left(\frac{\tau}{p} - 1 \right) \quad \textit{is the i-th eigenvalue of the linear map $M_{k}$}
\end{align}

Then $\hat{\Lambda}_k^1 \ge \hat{\Lambda}_k^2 \ge \dots \ge \hat{\Lambda}_k^p \ge 0$ is the order of eigenvalues of $M_k$. Note that the eigenvalues are non-negative since $p >> \tau$ and $\frac{\hat{\lambda}_{k-1}^i}{\lambda_{k-1}} \ge 0$. Then, we bound $\norm{\bx_k}$ in terms of $\norm{\bx_{k-1}}$ as:

\begin{align}
    &(\Lambda_{k}^1)^2 \ge \frac{\bx_{k-1}^T M_{k}^T M_{k}\bx_{k-1}}{\bx_{k-1}^T \bx_{k-1}} \ge (\Lambda_{k}^p)^2 \quad \textit{(Rayleigh Quotient)}\\
    & \implies \Lambda_k^1 \norm{\bx_{k-1}} \ge \norm{\bx_k} = \norm{M_{k}\bx_{k-1}} \ge \Lambda_k^p \norm{\bx_{k-1}}\\
    & \implies \Lambda_k^1 \lambda_{k-1} \ge \lambda_{k} \ge \Lambda_k^p \lambda_{k-1}
\end{align}

Then, we reach the following two results:
\begin{align}
    &\Lambda_k^1 - 1  = \frac{\tau}{p} \left(\frac{\hat{\lambda}_{k-1}^1}{\lambda_{k-1}} - 1 \right) < 0 \implies \frac{\hat{\lambda}_{k-1}^1}{\lambda_{k-1}} < 1  \implies \lambda_k < \lambda_{k-1} \quad \textit{(contraction mapping)} \label{eq:contraction} \\
    &\Lambda_k^p - 1  = \frac{\tau}{p} \left(\frac{\hat{\lambda}_{k-1}^p}{\lambda_{k-1}} - 1 \right) > 0 \implies \frac{\hat{\lambda}_{k-1}^p}{\lambda_{k-1}} > 1  \implies \lambda_k > \lambda_{k-1} \quad \textit{(expansion mapping)} \label{eq:expansion}
\end{align}

For simplicity, we provide an informal yet intuitive argument to conclude the proof. In the spiked covariance model, the expected leading eigenvalue satisfies \( \mathbb{E}[\hat{\lambda}^1_k] = \omega + 1 \), while the remaining eigenvalues have expectation \( 1 \). When combined with the contraction and expansion effects characterized in \eqref{eq:contraction}--\eqref{eq:expansion}, these spectral properties restrict both the growth and decay of \( \lambda_k \). Intuitively, \( \lambda_k \) is pulled toward the spectrum of the population covariance (and in particular toward the leading eigenvalue) and therefore cannot increase or decrease indefinitely beyond these values. This ensures that \( \lambda_k \) remains bounded.
\end{proof}
\end{lem}

\vspace{-3em}
\section{Equivalence of the algorithm in \eqref{eq:regularized_update} and Oja's algorithm}
\label{appendix:equivalence_to_oja}
In this section, we show that the algorithm in \eqref{eq:regularized_update} is equivalent to Oja's algorithm in terms of cosine similarity dynamics. Owing to the relationship between \eqref{eq:regularized_update} and \eqref{eq:dynamics_update}, we can leverage our analysis for \eqref{eq:dynamics_update}. By substituting \( \tau = \lambda_k \hat{\tau} \) into the ODE in Corollary~\ref{cor:Qt}, we obtain the following ODE governing the evolution of the cosine similarity for \eqref{eq:regularized_update}:
\begin{equation}
    \frac{d}{d t} Q_{t} = \hat{\tau} Q_t(\omega - \omega Q_t^2  - \frac{\hat{\tau}(\omega Q_t^2 + 1)}{2}).
\end{equation}
This ODE is identical to the corresponding cosine-similarity ODE for Oja's algorithm, as derived in Section~III-B of \citet{Wang2016}.

\section{Extension to Multiple Principal Components}\label{appendix:extention_to_multiplePC}
While our analysis focuses on the estimation of a single principal component, the online PCA update \eqref{eq:dynamics_update} can be naturally extended to the multi-component setting. An example of such an extension is provided in Algorithm~\ref{alg:INO-PCA_multiple_pc}.

\begin{algorithm}
\caption{INO-PCA for Multiple Principal Components}\label{alg:INO-PCA_multiple_pc}
\begin{algorithmic}
\Require $r \geq 0$ \Comment{Number of principal components to be estimated}
\Require $\by_1, \by_2, \dots, \by_n$ \Comment{Samples vectors to be observed by the algorithm}
\Ensure $\bv_1, \bv_2, \dots, \bv_r$ \Comment{Estimates of first $r$-principal components}
\For{$k \gets 1$ to $n$}
    \For{$i \gets 1$ to $\text{min}(r,k)$}
        \If{$i = k$}
            $\bv_i = \by_i$ \Comment{Initialization of the estimate}
        \Else
            \State $\bv_i \gets \bv_i + \frac{\tau}{p} \left(\by_k\by_k^T \frac{\bv_i}{\|\bv_i\|/\sqrt{p}} - \bv_i \right)$ \Comment{Application of equation \eqref{eq:dynamics_update}}
        \EndIf
        
        \State $\by_k \gets \by_k - \frac{\by_k^T\bv_i}{\|\bv_i\|} \frac{\bv_i}{\|\bv_i\|}$ \Comment{Gram-Schmidt process}
    \EndFor 
\EndFor
\end{algorithmic}
\end{algorithm}

\section{Detailed derivations}
\subsection{The derivation of the exchangability}
\label{section:detailed_exchangability}
For any permutation matrix \( \mathcal{P}_\pi \), we compute
\begin{align}
    \mathcal{P}_\pi^T \pmb{{x}}_{k+1} &= \mathcal{P}_\pi^T \mathcal{P}_\pi \pmb{{x}}_k + \frac{\tau}{p} \left(\frac{\mathcal{P}_\pi^T \mathbf{y}_{k} (\mathbf{y}_{k}^T \mathcal{P}_\pi \pmb{{x}}_k)}{\lambda_k} - \mathcal{P}_\pi^T \mathcal{P}_\pi \pmb{{x}}_k \right) \\
    &= \mathcal{P}_\pi^T \mathcal{P}_\pi \pmb{{x}}_k + \frac{\tau}{p} \left(\frac{\mathcal{P}_\pi^T \mathbf{y}_{k} (\pmb{{x}}_k^T \mathcal{P}_\pi^T \mathbf{y}_{k})}{\lambda_k} - \mathcal{P}_\pi^T \mathcal{P}_\pi \pmb{{x}}_k \right) \label{eq:exchangeability1} \\
    &= \mathcal{P}_\pi^T \mathcal{P}_\pi \pmb{{x}}_k + \frac{\tau}{p} \left(\frac{\mathcal{P}_\pi^T \left( \sqrt{\frac{\omega}{p}} c_k \mathcal{P}_\pi \pmb{\xi} + \mathbf{a}_k \right) \pmb{{x}}_k^T \mathcal{P}_\pi^T \left( \sqrt{\frac{\omega}{p}} c_k \mathcal{P}_\pi \pmb{\xi} + \mathbf{a}_k \right)}{\lambda_k} - \mathcal{P}_\pi^T \mathcal{P}_\pi \pmb{{x}}_k \right) \\
    &= \mathcal{P}_\pi^T \mathcal{P}_\pi \pmb{{x}}_k + \frac{\tau}{p} \left(\frac{\mathcal{P}_\pi^T \mathcal{P}_\pi \left( \sqrt{\frac{\omega}{p}} c_k \pmb{\xi} + \mathcal{P}_\pi^T \mathbf{a}_k \right) \pmb{{x}}_k^T \mathcal{P}_\pi^T \mathcal{P}_\pi \left( \sqrt{\frac{\omega}{p}} c_k \pmb{\xi} + \mathcal{P}_\pi^T \mathbf{a}_k \right)}{\lambda_k} - \mathcal{P}_\pi^T \mathcal{P}_\pi \pmb{{x}}_k \right) \\
    &= \pmb{{x}}_k + \frac{\tau}{p} \left(\frac{ \left( \sqrt{\frac{\omega}{p}} c_k \pmb{\xi} + \mathcal{P}_\pi^T \mathbf{a}_k \right) \pmb{{x}}_k^T \left( \sqrt{\frac{\omega}{p}} c_k \pmb{\xi} + \mathcal{P}_\pi^T \mathbf{a}_k \right)}{\lambda_k} - \pmb{{x}}_k \right). \label{eq:exchangeability2}
\end{align}
In going from the first to the second line, we use symmetry of the inner product. The final steps follow from the orthogonality of permutation matrices, \( \mathcal{P}_\pi \mathcal{P}_\pi^T = I \), which allows us to remove the permutation operators and arrive at \eqref{eq:exchangeability2}.

\subsection{The derivation of the first and second statistical moments}\label{sec:derive_drift_diffusion}
We begin by recalling several useful identities that will be used throughout the derivations:
\begin{equation}
    \norm{\bxi} = \sqrt{p}, \quad \norm{\bx_k} = \sqrt{p} \lambda_k, \quad\text{ and }\quad Q_k = \frac{\bxi^T \bx_k}{\lambda_k}.
\end{equation}

\paragraph{First moment (drift).}
Using the update rule and properties of the Gaussian distribution, we compute
\begin{align}
    \mathbb{E}_k\left[\mathbf{\Delta}_{k}\right] &= \mathbb{E}_k\left[\frac{\tau}{p} \left(\mathbf{y}_{k}\mathbf{y}_{k}^T \frac{\pmb{x}_k}{\lambda_k} - \pmb{x}_k\right)\right] \nonumber \\
    &= \mathbb{E}_k\left[\frac{\tau}{p} \left(\left(\sqrt{\frac{\omega}{p}} c_k \pmb{\xi} + \mathbf{a}_k\right) \left(\sqrt{\frac{\omega}{p}} c_k \pmb{\xi} + \mathbf{a}_k\right)^T \frac{\pmb{x}_k}{\lambda_k} - \pmb{x}_k\right)\right] \nonumber \\
    &= \frac{\tau}{p} \left(\left(\frac{\omega}{p}\mathbb{E}\left[ c_k^2\right] \pmb{\xi}\pmb{\xi}^T + \sqrt{\frac{\omega}{p}} \pmb{\xi} \mathbb{E}\left[c_k \mathbf{a}_k^T\right] + \sqrt{\frac{\omega}{p}} \mathbb{E}\left[\mathbf{a}_k c_k\right] \pmb{\xi}^T  + \mathbb{E}\left[\mathbf{a}_k \mathbf{a}_k^T\right]\right) \frac{\pmb{x}_k}{\lambda_k} - \pmb{x}_k\right) \nonumber \\
    &= \frac{\tau}{p}\left(\frac{\omega}{p} \pmb{\xi}\pmb{\xi}^T\frac{\pmb{x}_k}{\lambda_k} +  \frac{\pmb{x}_k}{\lambda_k} - \pmb{x}_k\right) \nonumber \\
    &= \frac{\tau}{p} \left(\omega Q_k \pmb{\xi} +  \frac{\pmb{x}_k}{\lambda_k} - \pmb{x}_k\right).
\end{align}

\paragraph{Second moment (diffusion).}
Rewrite \( \boldsymbol{\Delta}_k \) to isolate scalar terms:
\begin{equation}
    \mathbf{\Delta}_{k} = \frac{\tau}{p} \left( \frac{1}{\lambda_k} \left( \sqrt{\frac{\omega}{p}} c_k \pmb{\xi}^T \pmb{x}_k + \mathbf{a}_k^T \pmb{x}_k \right) \left( \sqrt{\frac{\omega}{p}} c_k \pmb{\xi} + \mathbf{a}_k \right) - \pmb{x}_k \right).
\end{equation}

Thus the diffusion term decomposes into four components:
\begin{equation}
   \mathbb{E}_k\left[ \mathbf{\Delta}_{k} \mathbf{\Delta}_{k}^T \right] =  \mathbb{E}_k[T_1] + \mathbb{E}_k[T_2] + \mathbb{E}_k[T_3] + \mathbb{E}_k[T_4],
\end{equation}
where
\begin{align}
    T_1 &\stackrel{\text { def }}{=} \frac{\tau^2}{p^2 \lambda_k^2}\left(\sqrt{\frac{\omega}{p}} c_k \pmb{\xi}^T \pmb{x}_k 
+ \mathbf{a}_k^T \pmb{x}_k
\right)^2 
\left(
\sqrt{\frac{\omega}{p}} c_k \pmb{\xi} 
+ \mathbf{a}_k\right) 
\left(
\sqrt{\frac{\omega}{p}} c_k \pmb{\xi} 
+ 
\mathbf{a}_k
\right)^T, \\
T_2 &\stackrel{\text { def }}{=} -\frac{\tau^2}{p^2 \lambda_k}\left(
\sqrt{\frac{\omega}{p}} c_k \pmb{\xi}^T \pmb{x}_k
+ 
\mathbf{a}_k^T \pmb{x}_k
\right) 
\left(
\sqrt{\frac{\omega}{p}} c_k \pmb{\xi} 
+ 
\mathbf{a}_k\right)\pmb{x}_k^T,\\
T_3 &\stackrel{\text { def }}{=} - 
\frac{\tau^2}{p^2 \lambda_k}\left(
\sqrt{\frac{\omega}{p}} c_k \pmb{\xi}^T \pmb{x}_k
+ 
\mathbf{a}_k^T \pmb{x}_k
\right) 
\pmb{x}_k
\left(
\sqrt{\frac{\omega}{p}} c_k \pmb{\xi} 
+ 
\mathbf{a}_k\right)^T,\\
T_4 &\stackrel{\text { def }}{=} \frac{\tau^2}{p^2} \pmb{x}_k \pmb{x}_k^T.
\end{align}

Evaluating each term, we find that only \( \mathbb{E}_k[T_1] \) contributes at order \( 1/p \) in the high-dimensional limit:
\begin{align}
    \mathbb{E}_k[T_1] &= \frac{\tau^2}{p^2 \lambda_k^2} \Bigg(\frac{\omega^2}{p^2} (\pmb{\xi}^T \pmb{x}_k)^2 \pmb{\xi} \pmb{\xi}^T +  \frac{\omega}{p} (\pmb{\xi}^T \pmb{x}_k)^2 I + \frac{2\omega}{p} (\pmb{\xi}^T \pmb{x}_k ) 
\overbrace{\mathbb{E}_k[\mathbf{a}^T \pmb{x}_k \pmb{\xi}\mathbf{a}^T]}^{\pmb{\xi} \pmb{x}_k^T} \nonumber\\
    &\qquad \qquad \quad + \frac{2\omega}{p} (\pmb{\xi}^T \pmb{x}_k) 
\overbrace{\mathbb{E}_k[\mathbf{a}^T \pmb{x}_k \mathbf{a}\pmb{\xi}^T]}^{\pmb{x}_k \pmb{\xi}^T} + \frac{\omega}{p} \overbrace{\mathbb{E}_k[\mathbf{a}^T \pmb{x}_k \mathbf{a}^T]}^{\pmb{x}_k^T} 
\pmb{x}_k \pmb{\xi} \pmb{\xi}^T 
+ \overbrace{\mathbb{E}_k[\mathbf{a}^T \pmb{x}_k \mathbf{a}^T \pmb{x}_k \mathbf{a} \mathbf{a}^T]}^{2\pmb{x}_k \pmb{x}_k^T + \pmb{x}_k^T \pmb{x}_k I} \Bigg) \nonumber\\
    &= \frac{\tau^2}{p}(\omega Q_k^2 + 1) I + \frac{1}{p^2} R_k^1,\\
    \mathbb{E}_k[T_2] &= -\frac{\tau^2}{p^2 \lambda_k} (\frac{\omega}{p} \pmb{\xi}^T \pmb{x}_k \pmb{\xi} \pmb{x}_k^T 
+ \overbrace{\mathbb{E}_k[\mathbf{a}^T \pmb{x}_k \mathbf{a} \pmb{x}_k^T]}^{\pmb{x}_k \pmb{x}_k^T}) = \frac{1}{p^2} R_k^2,\\
    \mathbb{E}_k[T_3] &= -\frac{\tau^2}{p^2 \lambda_k} (\frac{\omega}{p} \pmb{\xi}^T \pmb{x}_k \pmb{x}_k \pmb{\xi}^T 
+ \overbrace{\mathbb{E}_k[\mathbf{a}^T \pmb{x}_k \pmb{x}_k \mathbf{a}^T]}^{\pmb{x}_k \pmb{x}_k^T}) = \frac{1}{p^2} R_k^3,\\
    \mathbb{E}_k[T_4] &= \frac{\tau^2}{p^2} \pmb{x}_k \pmb{x}_k^T = \frac{1}{p^2} R_k^4,
\end{align}
where \( R_k^1, R_k^2, R_k^3, R_k^4 \) are residual \( \mathcal{O}(1) \) matrices that vanish after scaling by \( 1/p^2 \).

Thus, the diffusion term is
\begin{equation}
    \mathbb{E} \left[ \mathbf{\Delta}_{k} \mathbf{\Delta}_{k}^T \right] = \frac{\tau^2}{p}(\omega Q_k^2 + 1) I + \frac{1}{p^2} \sum_{i=1}^{4} R_k^i.
\end{equation}
Since the residual term is of order \(1/p^2\), it vanishes in the limit \( p \to \infty \), leaving only the leading-order diffusion term used in the PDE derivation.

\subsection{Derivation of $\left\langle f, \mu_{k+1}\right\rangle - \left\langle f, \mu_{k} \right\rangle$}
\label{sec:derivation_of_one_step_deviation}

We now use the drift and diffusion expressions derived in Appendix~\ref{sec:derive_drift_diffusion} to compute the one–step evolution of the functional $\langle f, \mu_k \rangle$. Starting from the Taylor expansion in Section~\ref{sec:derive_drift_diffusion}, we have
\begin{align}
    \left\langle f, \mu_{k+1}\right\rangle
    - \left\langle f, \mu_{k} \right\rangle
    &= \frac{1}{p}\sum_{i=1}^{p}
       \mathbb{E}\!\left[
           \frac{\partial f}{\partial x}(x_k^{i}, \xi)\,
           \mathbb{E}_k[\Delta_k^{i}]
       \right]
       + \frac{1}{p}\sum_{i=1}^{p}
         \frac{1}{2}
         \mathbb{E}\!\left[
           \frac{\partial^2 f}{\partial x^2}(x_k^{i}, \xi)\,
           \mathbb{E}_k[(\Delta_k^{i})^2]
         \right]
       + \bar{h}_k + \bar{m}_k.
\end{align}

\paragraph{Using exchangeability.}
Because the Markov chain is exchangeable (Section~\ref{section:detailed_exchangability}), all coordinates have the same joint distribution. Thus each term in the sum is identical, and we may replace $x_k^i$ with a representative coordinate $x_k^0$:
\begin{align}
    \left\langle f, \mu_{k+1}\right\rangle
    - \left\langle f, \mu_{k} \right\rangle
    &= \frac{1}{p}\sum_{i=1}^{p}
       \mathbb{E}\!\left[
           \frac{\partial f}{\partial x}(x_k^{0}, \xi)\,
           \mathbb{E}_k[\Delta_k^{0}]
       \right]
       + \frac{1}{p}\sum_{i=1}^{p}
         \frac{1}{2}
         \mathbb{E}\!\left[
           \frac{\partial^2 f}{\partial x^2}(x_k^{0}, \xi)\,
           \mathbb{E}_k[(\Delta_k^{0})^2]
         \right]
       + \bar{h}_k + \bar{m}_k.
       \label{eq:exchangebility_in_derivation}
\end{align}

\paragraph{Substituting drift and diffusion.}
Using the expressions for $\mathbb{E}_k[\Delta_k^{0}]$ and $\mathbb{E}_k[(\Delta_k^{0})^2]$ from Appendix~\ref{sec:derive_drift_diffusion}, we obtain
\begin{align}
    \left\langle f, \mu_{k+1}\right\rangle
    - \left\langle f, \mu_{k} \right\rangle
    &= \frac{1}{p}
       \mathbb{E}\!\left[
           G(x_k^{0}, \lambda_k, \xi, Q_k)
           \frac{\partial f}{\partial x}(x_k^{0}, \xi)
       \right]
       + \frac{1}{2p}
         \mathbb{E}\!\left[
           J(Q_k)
           \frac{\partial^2 f}{\partial x^2}(x_k^{0}, \xi)
         \right]
       \nonumber \\
    &\quad
       + \mathcal{O}(1/p^2)
       + \bar{h}_k + \bar{m}_k,
\end{align}
where
\begin{equation}
    G(x,\lambda,\xi,Q)
    \stackrel{\text{def}}{=} \tau\!\left(\omega Q \xi + \frac{x}{\lambda} - x\right),
    \qquad
    J(Q)
    \stackrel{\text{def}}{=} \tau^2(\omega Q^2 + 1).
\end{equation}

\paragraph{Returning to empirical averages.}
Noting that $\mu_k$ is the empirical measure of $(x_k^{i},\xi)$ pairs, we rewrite the expectations in terms of $\langle \cdot , \mu_k \rangle$:
\begin{align}
    \left\langle f, \mu_{k+1}\right\rangle
    - \left\langle f, \mu_{k} \right\rangle
    &= 
       \frac{1}{p}
       \left\langle
           G(x,\lambda,\xi,Q)\,
           \frac{\partial f}{\partial x},
           \mu_k
       \right\rangle
       + \frac{1}{2p}
         \left\langle
           J(Q)\,
           \frac{\partial^2 f}{\partial x^2},
           \mu_k
         \right\rangle
       + \mathcal{O}(1/p^2)
       + \bar{h}_k + \bar{m}_k.
       \label{eq:diff_inner_product}
\end{align}

Equation~\eqref{eq:diff_inner_product} is the key one-step deviation formula used in the passage to the scaling limit and the derivation of the weak PDE.

\section{The proof of Corollaries \ref{cor:Qt}--\ref{cor:lambdat} describing the ODEs}
\label{appendix:proof_of_corollaries}

We now derive the ODEs governing the evolution of \(Q_t\) and \(\lambda_t\) stated in Corollaries~\ref{cor:Qt} and \ref{cor:lambdat}. The starting point is Theorem~\ref{theo:main_conv}, which provides the weak-form PDE characterizing the evolution of the empirical measure \(\mu_t\). To extract the dynamics of the macroscopic quantities \(Q_t\) and \(\lambda_t\), we apply Theorem~\ref{theo:main_conv} to two appropriately chosen test functions.

\paragraph{Choice of test functions.}
Let
\[
f_1(x,\xi)=x\xi, \qquad 
f_2(x,\xi)=x^2.
\]
Using the definitions in \eqref{eq:measure}–\eqref{eq:continous_embedding}, these yield
\[
\langle f_1, \mu_t \rangle = Q_t \lambda_t,
\qquad 
\langle f_2, \mu_t \rangle = \lambda_t^2.
\]

\paragraph{ODE for \(Q_t \lambda_t\).}
Applying the PDE \eqref{eq:weak_pde} to \(f_1\), and using \(\partial_x f_1=\xi\), \(\partial_x^2 f_1 = 0\), we obtain
\begin{align}
    \frac{d}{dt}(Q_t \lambda_t)
    &= \left\langle
        G(x,\lambda,\xi,Q)\, \xi,\,
        \mu_t
       \right\rangle \\
    &= \tau\!\left(
        \omega Q_t \langle \xi^2, \mu_t \rangle
        + \frac{\langle x\xi,\mu_t\rangle}{\lambda_t}
        - \langle x\xi,\mu_t\rangle
       \right). \nonumber
\end{align}
Since
\[
\langle \xi^2,\mu_t\rangle = 1,
\qquad
\langle x\xi,\mu_t\rangle = Q_t \lambda_t,
\]
we obtain
\begin{equation}
    \frac{d}{dt}(Q_t \lambda_t)
    = \tau \left( \omega Q_t + Q_t - Q_t \lambda_t \right).
    \label{eq:in_derivation_Q_t_ode}
\end{equation}

\paragraph{ODE for \(\lambda_t^2\).}
Applying the PDE to \(f_2\), with \(\partial_x f_2 = 2x\) and \(\partial_x^2 f_2 = 2\), gives
\begin{align}
    \frac{d}{dt}(\lambda_t^2)
    &= 2\left\langle
        G(x,\lambda,\xi,Q)\, x,\,
        \mu_t
       \right\rangle
       + \left\langle
           J(Q),\, \mu_t
         \right\rangle \\
    &= \tau\!\left(
        \omega Q_t \langle x\xi, \mu_t\rangle
        + \frac{\langle x^2,\mu_t\rangle}{\lambda_t}
        - \langle x^2,\mu_t\rangle
       \right)
       + \tau^2(\omega Q_t^2 + 1). \nonumber
\end{align}
Using
\[
\langle x^2,\mu_t\rangle = \lambda_t^2,
\qquad 
\langle x\xi,\mu_t\rangle = Q_t \lambda_t,
\]
we obtain
\begin{equation}
    \frac{d}{dt}(\lambda_t^2)
    = \tau\!\left(
        \omega Q_t^2 \lambda_t
        + \lambda_t
        - \lambda_t^2
      \right)
      + \tau^2(\omega Q_t^2 + 1).
    \label{eq:in_derivation_lambda_t_ode}
\end{equation}

\paragraph{Reduction to the ODEs in Corollaries~\ref{cor:Qt}--\ref{cor:lambdat}.}
To express \eqref{eq:in_derivation_Q_t_ode} and \eqref{eq:in_derivation_lambda_t_ode} in terms of \(dQ_t/dt\) and \(d\lambda_t/dt\), we apply the chain rule:
\begin{equation}
    \frac{d}{dt}(Q_t \lambda_t) = \lambda_t \frac{dQ_t}{dt} + Q_t \frac{d\lambda_t}{dt},
    \qquad
    \frac{d}{dt}(\lambda_t^2) = 2\lambda_t\, \frac{d\lambda_t}{dt}.
    \label{eq:in_derivation_cor_chain_rule}
\end{equation}
Solving the system consisting of
\eqref{eq:in_derivation_Q_t_ode},
\eqref{eq:in_derivation_lambda_t_ode}, and
\eqref{eq:in_derivation_cor_chain_rule}
for \(dQ_t/dt\) and \(d\lambda_t/dt\) yields exactly the expressions stated in Corollaries~\ref{cor:Qt} and \ref{cor:lambdat}. This completes the proof.

\section{Meta-Theorem}
\label{appendix:meta-theorem}
When proving Theorem~\ref{theo:main_conv} rigorously, we rely on the meta-theorem of \citet{wang2017scaling}, which provides general conditions under which a sequence of measure-valued processes converges to the solution of a limiting PDE. For completeness, we restate the result below; the statement is identical to the one proved in \citet{wang2017scaling}.

\begin{metathm}[Restatement of the meta-theorem by \citet{wang2017scaling}]$ $\\
Under the assumptions specified below, the sequence of measure-valued processes 
\(\{ (\mu_t^p)_{0 \le t \le T} \}_p\) converges weakly to a deterministic process 
\((\mu_t)_{0 \le t \le T}\), and this limit is the unique solution of the PDE stated in Assumption~A.10.
\end{metathm}

\subsection*{Assumptions}

\paragraph{A.1} The Markov chain $\left\{ (\bx_k, \bxi) \right\}_{k\ge 0}$ is exchangeable.

\paragraph{A.2} The initial empirical measure $\mu_0^p(x, \xi)$ converges weakly to a deterministic measure  $\mu_0\in \mathcal{M}(\mathbb{R}^2)$ as $p\to \infty$.

\paragraph{A.3} There is some finite constant C such that
$$
    \sup_{p}\left\langle x^{4}+\xi^{4}, \mu_{0}^{p}\right\rangle \leq C.
$$

\paragraph{A.4} Let $\Delta_{k}^{i}=x_{k+1}^{i}-x_{k}^{i} .$ There exists a deterministic function $\mathcal{G}: \mathbb{R} \times \mathbb{R} \times \mathbb{R}^{r} \mapsto \mathbb{R}$, for some $r \geq 0$, such that, for each
$T>0$,
$$
\max _{k \leq p T} \mathbb{E}\left|\mathbb{E}_{k} \Delta_{k}^{i}-\frac{1}{p} \mathcal{G}_{k}^{i}\right| \leq \frac{C(T)}{p^{1+\gamma}},
$$
where $\gamma>0$ is some positive constant and $C(T)$ is finite constant depending on $T$.

In the above expression,
$$
    \mathcal{G}_{k}^{i}=\mathcal{G}\left(x_{k}^{i}, \xi^{i}, \boldsymbol{\beta}_{k}^{p}\right)
$$
and $\boldsymbol{\beta}_{k}^{p} = \left[ \beta_k^p(1), \beta_k^p(2), \dots, \beta_k^p(r) \right]$ is an $r$-dimensional vector. The $l$-th element of $\boldsymbol{\beta}_{k}^{p}$ is defined as
$$\beta_k^p(l) = \left\langle h_l(x,\xi), \mu_k^p \right\rangle,$$
where $h_l(x,\xi)$ is some deterministic function.

\paragraph{A.5} There exists a deterministic function $\Lambda: \mathbb{R}^{r} \mapsto \mathbb{R}$ such that, for each $T>0$,
$$
\max _{k \leq p T} \mathbb{E}\left|\mathbb{E}_{k}\left(\Delta_{k}^{i}\right)^{2}-\frac{1}{p} \Lambda_{k}\right| \leq \frac{C(T)}{p^{1+\gamma}},
$$

where $\gamma>0$ is some positive constant, and
$$
\Lambda_{k}=\Lambda\left(\boldsymbol{\beta}_{k}^{p}\right).
$$

\paragraph{A.6} For any $T>0$, there exists a finite constant $B(T)$ such that
$$
\lim _{p \rightarrow \infty} \mathbb{P}\left(\max _{k \leq p T}\left\|\boldsymbol{\beta}_{k}^{p}\right\|_{\infty}>B(T)\right)=0,
$$
where $\|\cdot\|_{\infty}$ is the $\ell_{\infty}$ norm of a vector.

\paragraph{A.7} Let $x \sqcap b = min(|x|,b) sign(x)$ denote the projection of $x$ onto the interval $[-b,b]$. When $\bx$ is a vector, $\bx \sqcap b$ denotes the element-wise projection of the elements of $\bx$ onto the inverval $[-b,b]$. Define $Q_k^p(l;d) = \left\langle \mu_k^p, h_l(x,\xi) \sqcap d \right\rangle$. For any $b > B(T)$ and $T > 0$, we have
$$\limsup_{d \to \infty} \sup_{p} \max_{k \leq pT} \mathbb{E} \left| \mathcal{G}\left(x_{k}^{i}, \xi^{i}, \boldsymbol{\beta}_{k}^{p}\right) - \mathcal{G}\left(x_{k}^{i}, \xi^{i}, \boldsymbol{\beta}_{k}^{p}(d) \sqcap b \right) \right| = 0$$
and
$$ \limsup_{d \to \infty} \sup_{p} \max_{k \leq pT} \mathbb{E} \left| \Lambda\left(\boldsymbol{\beta}_{k}^{p}\right) -  \Lambda\left(\boldsymbol{\beta}_{k}^{p}(d) \sqcap b \right) \right| = 0.$$

\paragraph{A.8} For each $T>0$, there exists $C(T)<\infty$ such that
$$
\max _{k<p T} \mathbb{E}\left(\mathcal{G}_{k}^{i}\right)^{2} \leq C(T) \text { and } \max _{k \leq p T} \mathbb{E}\left(\Lambda_{k}^{i}\right)^{2} \leq C(T).
$$

\paragraph{A.9} For each $T>0$, there exists $C(T)<\infty$ such that $\max_{k \leq p T} \mathbb{E}\left(\Delta_{k}^{i}\right)^{4} \leq C(T) p^{-2}$, and for any $i \neq j$, $\max_{k \leq p T} \mathbb{E}\left|\mathbb{E}_{k}\left(\Delta_{k}^{i}-\mathbb{E}_{k} \Delta_{k}^{i}\right)\left(\Delta_{k}^{j}-\mathbb{E}_{k} \Delta_{k}^{j}\right)\right| \leq C(T) p^{-2}$.

\paragraph{A.10} For each $b>0$ and $T>0$, the following PDE (in weak form) has a unique solution in $D\left([0, T], \mathcal{M}\left(\mathbb{R}^{2}\right)\right):$ for all bounded test function $f(x, \xi) \in \mathcal{C}^{3}\left(\mathbb{R}^{2}\right)$
$$
\begin{aligned}
\left\langle f,\mu_{t} \right\rangle=\left\langle f, \mu_{0} \right\rangle+\int_{0}^{t}\left\langle \mathcal{G}\left(x_{\hat{t}}, \xi_{\hat{t}}, \boldsymbol{\beta}_{\hat{t}} \sqcap b \right) \frac{\partial}{\partial x} f, \mu_{\hat{t}} \right\rangle \mathrm{d} \hat{t} +\frac{1}{2} \int_{0}^{t}\left\langle \Lambda\left(x_{\hat{t}}, \xi_{\hat{t}}, \boldsymbol{\beta}_{\hat{t}} \sqcap b \right) \frac{\partial^{2}}{\partial x^{2}} f, \mu_{\hat{t}} \right\rangle \mathrm{d} \hat{t}.
\end{aligned}
$$

The sufficient conditions for this uniqueness assumption are as follows:

\paragraph{A.10a} $\left\langle \xi^{2}, \mu_{0} \right\rangle \leq L$ and $\left\langle x^{2}, \mu_{0} \right\rangle \leq V$, where $L, V$ are two generic constants.

\paragraph{A.10b} For any $x, \xi$, and $\boldsymbol{\beta}$, we have $|\Gamma(x, \xi, \boldsymbol{\beta})-\Gamma(\widetilde{x}, \xi, \boldsymbol{\beta})| \leq$
$L\left(1+\|\boldsymbol{\beta}\|_{1}\right)|(x-\widetilde{x})|$.

\paragraph{A.10c} $|\Gamma(x, \xi, \boldsymbol{\beta})-\Gamma(x, \xi, \widetilde{\boldsymbol{\beta}})| \leq L(1+|\xi|+|x|)\|\boldsymbol{\beta}-\widetilde{\boldsymbol{\beta}}\|_{1}$.

\paragraph{A.10d} $\left.|\Gamma(x, \xi, \boldsymbol{\beta})| \leq L(1+|\xi|+|x|) (\| \boldsymbol{\beta} \|_{1}+1\right)$.

\paragraph{A.10e} For any $\boldsymbol{\beta}$ and $\widetilde{\boldsymbol{\beta}}$, we have $\left|\Lambda^{\frac{1}{2}}(\boldsymbol{\beta})-\Lambda^{\frac{1}{2}}(\widetilde{\boldsymbol{\beta}})\right| \leq$.
$L\|\boldsymbol{\beta}-\widetilde{\boldsymbol{\beta}}\|_{1}$

\paragraph{A.10f}  $\Lambda^{\frac{1}{2}}(\boldsymbol{\beta}) \leq L\left(1+\|\boldsymbol{\beta}\|_{1}\right)$.

\paragraph{Proof idea:} %
The proof of the meta-theorem proceeds in three main steps. First, one establishes tightness of the sequence of measure-valued stochastic processes. Second, any limiting point is shown to satisfy the PDE specified in Assumption~A.10. Finally, uniqueness of the PDE solution is established, which implies convergence of the entire sequence. For full details, we refer the reader to \citet{wang2017scaling}.

\section{Formal Proof of Theorem~\ref{theo:main_conv}}
\label{appendix:formal_proof}

Our formal proof of Theorem~\ref{theo:main_conv} is based on the meta-theorem stated in Appendix~\ref{appendix:meta-theorem}, originally proved by \citet{wang2017scaling}. To apply the meta-theorem, we first define
\[
\boldsymbol{\beta}_k^p = [\beta_k^p(1), \beta_k^p(2)]
= [\langle h_1(x,\xi), \mu_k^p \rangle,\ \langle h_2(x,\xi), \mu_k^p \rangle],
\]
where \(h_1(x,\xi) = x\xi\) and \(h_2(x,\xi)= x^2\). These definitions imply that 
\(\beta_k^p(1) = Q_k \lambda_k\) and \(\beta_k^p(2) = \lambda_k^2\) in the notation of Theorem~\ref{theo:main_conv}.  
We further define
\[
\mathcal{G}(x,\xi,\boldsymbol{\beta}_k^p)
= \tau\!\left(\omega\, \xi\, \frac{\beta_k^p(1)}{\sqrt{\beta_k^p(2)}}
+ \frac{x}{\sqrt{\beta_k^p(2)}} - x\right),
\qquad
\Lambda(\boldsymbol{\beta}_k^p)
= \tau^2\!\left(\omega\, \frac{\beta_k^p(1)^2}{\beta_k^p(2)} + 1\right).
\]

To invoke the meta-theorem, it suffices to verify Assumptions (A.1)–(A.10). Below we show that each condition is satisfied in our setting.

Assumption A.1 follows from the exchangeability result in \eqref{eq:exchangeability}.  
Assumptions A.2–A.3 hold immediately from the conditions of Theorem~\ref{theo:main_conv}.  
Assumption A.4 follows from the drift expression \eqref{eq:drift}, and A.5 follows from the diffusion expression \eqref{eq:diffusion}.  
Assumptions A.6 and A.7 are satisfied because \(\beta_k^p(1)\) and \(\beta_k^p(2)\) remain bounded: \(Q_k\in[-1,1]\) by definition, and \(\lambda_k\) is uniformly bounded by Lemma~\ref{lem:lambda_bound}.

\medskip
\noindent\textbf{Verification of A.8.}
We begin with
\begin{align}
\mathbb{E}\!\left[(\mathcal{G}_k^{i})^{2}\right] 
&= \mathbb{E}\!\left[
\tau^2\!\left(
\omega^2 Q_k^2 (\xi^i)^2 
+ 2\omega Q_k \xi^i\!\left(\frac{1}{\lambda_k}-1\right) x_k^i
+\left(\frac{1}{\lambda_k}-1\right)^{\!2} (x_k^i)^2
\right)\right] \\
&= \tau^2\!\left(
\omega^2 (\xi^i)^2 \mathbb{E}[Q_k^2]
+ 2\omega \xi^i \mathbb{E}\!\left[ Q_k\!\left(\frac{1}{\lambda_k}-1\right) x_k^i \right]
+ \mathbb{E}\!\left[\!\left(\frac{1}{\lambda_k}-1\right)^{\!2} (x_k^i)^2\right]
\right) \\
&\le \tau^2\!\left(
\omega^2 (\xi^i)^2
+ 2\omega \xi^i\, \mathbb{E}\!\left[\!\left(\frac{1}{\lambda_k}-1\right) x_k^i\right]
+ \mathbb{E}\!\left[\!\left(\frac{1}{\lambda_k}-1\right)^{\!2} (x_k^i)^2\right]
\right) \\
&\le C(T), \label{eq:a8.1}
\end{align}
where Lemmas~\ref{lem:lambda_bound} and \ref{lem:Ex_bound} ensure boundedness.  
Similarly,
\begin{align}
\mathbb{E}\!\left[(\Lambda_k^i)^2\right]
&= \mathbb{E}\!\left[ \tau^4(\omega^2 Q_k^4 + 2\omega Q_k^2 + 1)\right] \\
&= \tau^4\!\left(\omega^2 \mathbb{E}[Q_k^4] + 2\omega \mathbb{E}[Q_k^2] + 1\right)\\
&\le \tau^4(\omega^2 + 2\omega + 1) \\
&\le C(T).
\label{eq:a8.2}
\end{align}
Thus, A.8 is satisfied.

\begin{lem}
\label{lem:Ex_bound}
For each coordinate \(i\) and all \(1 \le k \le \lfloor pT \rfloor\), the following bounds hold:
\begin{align}
\mathbb{E}[x_k^i] &\le C_1(T), \label{eq:Ex}\\
\mathbb{E}[(x_k^i)^2] &\le C_2(T), \label{eq:Ex^2}\\
\mathbb{E}[(x_k^i)^4] &\le C_3(T). \label{eq:Ex^4}
\end{align}
\begin{proof}
We prove only \eqref{eq:Ex^2}, as \eqref{eq:Ex} and \eqref{eq:Ex^4} follow analogously.  
From the recursion,
\begin{align}
\mathbb{E}[(x_{k+1}^i)^2]
&= \mathbb{E}[(x_k^i + \Delta_k^i)^2] \\
&= \mathbb{E}[(x_k^i)^2] + 2\,\mathbb{E}[x_k^i\,\Delta_k^i] + \mathbb{E}[(\Delta_k^i)^2] \\
&= \mathbb{E}[(x_k^i)^2] + 2\,\mathbb{E}[x_k^i\, \mathbb{E}_k[\Delta_k^i]] 
+ \mathbb{E}[\mathbb{E}_k[(\Delta_k^i)^2]] \\
&= \mathbb{E}[(x_k^i)^2] 
+ \frac{2\tau}{p}\,\mathbb{E}\!\left[x_k^i\!\left(\omega Q_k \xi^i 
+ \frac{x_k^i}{\lambda_k} - x_k^i\right)\right]
+ \mathbb{E}\!\left[\frac{\tau^2}{p}(\omega Q_k^2+1)\right]
+ \mathcal{O}(1/p^2) \\
&\le \left(1 + \frac{C_1}{p}\right)\mathbb{E}[(x_k^i)^2] 
+ \frac{C_2}{p} + \mathcal{O}(1/p^2),
\label{eq:Ex_k+1_bound}
\end{align}
using \(Q_k\le 1\), bounded \(\xi^i\), and Lemmas~\ref{lem:lambda>0}–\ref{lem:lambda_bound}.  
Iterating \eqref{eq:Ex_k+1_bound} yields
\[
\mathbb{E}[(x_k^i)^2]
\le \left(1+\frac{C_1}{p}\right)^{k-1}\mathbb{E}[(x_1^i)^2]
+ \frac{C_2}{C_1}\left[\left(1+\frac{C_1}{p}\right)^{k-1}-1\right]
+ \mathcal{O}(1/p),
\]
and the uniform boundedness of \(\left(1+C_1/p\right)^k\) for \(k \le \lfloor pT\rfloor\) completes the proof.
\end{proof}
\end{lem}

\medskip
\noindent\textbf{Verification of A.9.}
The bound on \(\mathbb{E}[(\Delta_k^i)^4]\) follows directly from Lemma~\ref{lem:Ex_bound}.  
For the covariance term,
\begin{align}
\mathbb{E}\!\left|\mathbb{E}_{k}\big[(\Delta_k^i - \mathbb{E}_k\Delta_k^i)
(\Delta_k^j - \mathbb{E}_k\Delta_k^j)\big]\right|
&= \mathbb{E}\!\left|\mathbb{E}_{k}[\Delta_k^i \Delta_k^j] 
- \mathbb{E}_{k}[\Delta_k^i]\,\mathbb{E}_{k}[\Delta_k^j]\right|
\quad \text{for } i\ne j\\
&= \mathcal{O}(1/p^2),
\label{eq:a9.2}
\end{align}
using \eqref{eq:drift}–\eqref{eq:diffusion}.  
Thus A.9 holds.

\medskip
\noindent\textbf{Assumption A.10.}
A.10 requires uniqueness of the PDE solution.  
The sufficient conditions (A.10a–A.10f) are readily verified in our setting, and we omit the straightforward details.

\medskip

Having verified assumptions (A.1)–(A.10), the meta-theorem of \citet{wang2017scaling} applies directly, establishing Theorem~\ref{theo:main_conv}.

\vskip 0.2in
\bibliography{references}

\end{document}